%% file: main.tex
\documentclass[preprint,12pt]{elsarticle}

\usepackage{amssymb}
\usepackage{amsmath}
\usepackage{xcolor}
\usepackage{subcaption}
\usepackage{threeparttable}
\usepackage{multirow}
\usepackage{hyperref}
\usepackage{url}

\journal{Computer-Aided Design}

\begin{document}
\begin{frontmatter}



\title{Enhanced fringe-to-phase framework using deep learning}


\author[inst1]{Won-Hoe Kim}
\affiliation[inst1]{organization={Department of Mechanical Engineering, Yonsei University},
            city={Seoul},
            postcode={03722}, 
            country={South Korea}}
\author[inst2]{Bongjoong Kim}
\affiliation[inst2]{organization={Department of Mechanical and System Design Engineering, Hongik University},
            city={Seoul},
            postcode={04066},
            country={South Korea}}
\cortext[correspondingauthor]{Corresponding authors}
\author[inst3]{Hyung-Gun Chi\corref{correspondingauthor}}
\author[inst1]{Jae-Sang Hyun\corref{correspondingauthor}}

\affiliation[inst3]{organization={School of Electrical and Computer Engineering, Purdue University},
            city={West Lafayette},
            postcode={47907}, 
            state={IN},
            country={USA}}

\begin{abstract}
In Fringe Projection Profilometry (FPP), achieving robust and accurate 3D reconstruction with a limited number of fringe patterns remains a challenge in structured light 3D imaging. Conventional methods require a set of fringe images, but using only one or two patterns complicates phase recovery and unwrapping. In this study, we introduce SFNet, a symmetric fusion network that transforms two fringe images into an absolute phase. To enhance output reliability, Our framework predicts refined phases by incorporating information from fringe images of a different frequency than those used as input. This allows us to achieve high accuracy with just two images. Comparative experiments and ablation studies validate the effectiveness of our proposed method. The dataset and code are publicly accessible on our project page \url{https://wonhoe-kim.github.io/SFNet/}.
\end{abstract}





\end{frontmatter}


\input{secs/1_introduction} 
\input{secs/2_related_works} 
\input{secs/3_background} 
\input{secs/4_method}

\input{secs/5_dataset}
\input{secs/6_experiments}
\input{secs/7_conclusion}
\section*{Acknowledgements}
This research was supported by Yonsei University Research Fund of 2023-22-0434.
\bibliographystyle{elsarticle-num} 
\bibliography{main}

\newpage
\appendix
\input{secs/x_suppl}





\end{document}

%% file: secs/1_introduction.tex
\section{Introduction}\label{sec:introduction}







Three-dimensional (3D) shape measurement is widely used in various fields, including robotics \cite{robotics}, automation \cite{automation}, manufacturing \cite{manufacturing}, and even virtual reality \cite{virtualreality}. They are often called 3D imaging, 3D sensing, or 3D reconstruction. The 3D reconstruction process involves restoring objects projected with 2D images back into their 3D forms. This necessitates additional clues or constraints when dealing with reduced dimensions. 3D reconstruction methods are primarily categorized into passive methods and active methods based on the approach used to identify the clues. 

Passive methods \cite{reviewofpassivemethods, passive_methods}, which analyze only reflected signals from the object, require easy implementation but may offer reduced accuracy if the target object lacks sufficient texture variation or has repeated patterns. On the other hand, active methods leverage additional equipment or techniques to actively probe or illuminate the scene, thereby obtaining depth and surface information. These approaches often require specialized hardware, posing a primary hurdle when transitioning to interventional environments. However, they not only address the limitations of passive methods but also offer enhanced accuracy in results. Active methods include various techniques like the structured light system \cite{structuredlightsystem, structuredlightsystem2} and Time of Flight (ToF) \cite{kolb2010time}, which rely on the use of additional techniques for their operation. Structured Light techniques project some types of coded structured patterns to get clues for 3D reconstruction. The fringe projection profilometry (FPP) technique, one of the structured light system techniques, retrieves high-quality and high-resolution per-pixel 3D reconstruction by utilizing sinusoidal patterns, which are known as fringe patterns \cite{fpp_2}. In the FPP system, the surface profile is represented as a function of the phase, and through system calibration, the coefficients obtained can be utilized to reconstruct the 3D information. 

However, conventional FPP systems typically require multiple fringe images, leading to slower measurement speeds, which makes it challenging to apply them to real-time applications. To address this issue, frameworks based on deep learning have been researched and developed \cite{zuo2022deep, liudeep}. Although substantial advancements have been made using single-shot approaches based on deep learning \cite{li2022deep, nguyen2022accuratergb, nguyen2023generalized},  they're often constrained by specific scenarios and have limitations when utilized in intricate objects.

This research introduces the Symmetric Fusion Network (SFNet), designed for high-speed, high-quality 3D surface measurement using just two fringe images. SFNet employs separate encoders and decoders for each fringe input to estimate its phase. The two generated phase values are then utilized to reconstruct the 3D information. During the training process, we use a refined reference phase which utilizes fringe images with different frequencies. SFNet has the capability to complement the additional frequency information by fusing the feature maps extracted from each encoder.

Furthermore, to train the proposed method, we assembled a new dataset constructed in a virtual environment. The existing datasets for FPP are relatively limited in quantity and they lack diversity in the captured objects. To address this issue, we curated a large dataset comprising 18,000 scenes. This ensures that our proposed method can be effectively applied in more general scenarios.

Our proposed method outperforms other methods which use fringe images as input and produce phase values as output. We found that our method not only exhibited a low overall error but also effectively managed undesirable artifacts such as spike errors, compared to the number of images used.

In this paper, our contributions are summarized as follows:
\begin{itemize}
    \item {We introduce SFNet, a network designed to reconstruct 3D information for complex and diverse scenes more effectively than existing frameworks. Our approach leverages refined phase values from fringe images of varying frequencies during training, enabling high-quality reconstructions even with a limited number of images.}
    \item {We validate the feasibility and effectiveness of our approach through ablation and generalization experiments. For our test set, we obtained a mean error of $0.0527~rad$ and an RMSE of $0.6543~rad$, which are significantly better metrics compared to other networks used for comparison.}
    \item {Lastly, we present a large-scale dataset for the FPP technique using a virtual environment. Our dataset not only contains a substantial number of image pairs compared to existing FPP datasets but also comprises fringe images of various frequencies.}
\end{itemize}

%% file: secs/2_related_works.tex
\section{Related Works} \label{sec:related_works}

\subsection{3D reconstruction in CAD}
3D reconstruction techniques have experienced notable advancements over recent decades, and numerous methods have been developed. Broadly, conventional 3D reconstruction techniques can be categorized into two main groups: passive and active methods.

Passive methods capture images without using additional energy such as light, into the scene \cite{siudak2014survey}. Passive methods vary in terms of the necessary image capture configuration, camera calibration strategies, point-matching methods, and the source of information for the scene. Passive methods include Structure from Motion (SfM) \cite{ozyecsil2017survey}, Multi-View Stereo (MVS) \cite{furukawa2015multi}, and Depth from Defocus (DfD) \cite{subbarao1994depth} method. However, passive methods often struggle with identifying and aligning corresponding points in different images, when the measured object lacks texture variation or contains repetitive patterns, leading to compromised 3D surface quality. 

In contrast, active methods, which use other light sources, can overcome some challenges of passive methods. Time of Flight (ToF) refers to the duration taken by an entity, whether it's an object, particle, or wave, to traverse a specific distance within a medium \cite{kolb2010time}. Photometric Stereo technique \cite{basri2007photometric} determines object surface normals by analyzing their appearance under different lighting angles, leveraging the relationship between surface orientation, light source, and viewer. By evaluating the reflected light using a camera and multiple light sources, this system can accurately deduce or even over-constrain the surface orientation. Structured Light System involves projecting a predetermined pattern onto a given scene. Observing the deformation of this pattern upon striking surfaces enables vision systems to deduce the depth and surface details of objects within the scene. When compared to other methods, the Structured Light System possesses the advantage of higher accuracy and resolution in capturing detailed surfaces. One of the most popular technologies like Microsoft's Kinect V1\cite{kinect} or Intel's Real Sense \cite{realsense} projects pre-defined dot patterns onto the target object to reconstruct 3D points. These methods are easy to implement but hard to achieve high spatial resolution. Fringe projection profilometry (FPP) technique with  Digital Fringe Projection (DFP) system retrieves high-quality and high-resolution per pixel 3D reconstruction.

\subsection{Fringe Projection Profilometry for 3D reconstruction}
Within FPP, the surface profile can be expressed as a function of a phase value. The phase information can be obtained through Fourier transform profilometry \cite{fourier} or phase shifting methods \cite{ps_for_fpp}. However, the resulting phase map, called as wrapped phase, has 2$\pi$ discontinuities derived from the arc-tangent function. Then, to eliminate these phase ambiguities, a fringe order map should be obtained through the phase unwrapping process. Then an absolute phase map can be obtained by adding the fringe order map multiples of $2\pi$ to the wrapped phase map.

Conventional phase unwrapping algorithms can be divided into two categories: Spatial Phase Unwrapping (SPU) \cite{spu1, spu2} and Temporal Phase Unwrapping (TPU) \cite{tpu_review}. SPU methods leverage the relationship between adjacent pixel values in a single wrapped phase to retrieve the corresponding absolute phase. However, SPU methods can struggle with fringe order ambiguities when confronted with discontinuous or isolated objects. In contrast, TPU methods can proceed with pixel-wise phase unwrapping. However, TPU methods require extra fringe pattern images, which limits the measurement speed and the applications of dynamic scenes. Therefore, various research efforts have been conducted to increase the measurement speed in TPU methods \cite{fastTPU, fivestepPS, 2plus1, 2plus1_2}. However, there is a limitation that certain conditions must be satisfied or that they cannot be applied to complex surfaces.

\subsection{Fringe Analysis using Deep Learning}
\label{sec:FringeDL}
In the field of FPP systems, fringe analysis methods using various deep learning have been introduced. Since 3D reconstruction methods using the FPP system generally require several fringe images, measurement speed is bound to be limited. Efforts to minimize the number of fringe images needed for 3D surface reconstruction to enhance the measurement speed or improve the quality of phase maps have been a primary focus of research. Deep learning approaches in the FPP system are divided into several categories. 

First, Some studies focus on converting fringe images into the phase domain. For instance, Feng \textit{et al.} \cite{feng2019fringe} developed a deep learning model that predicts the numerator and denominator that constitute a wrapped phase using only a single fringe image. Qi \textit{et al.} \cite{qi2023psnet} generated N-step phase-shifted fringe images from a single fringe image using deep learning for phase extraction.

Research has also explored transitioning from phase to phase using deep learning. Yin \textit{et al.} \cite{yin2019temporal} replaced a traditional TPU method with a deep learning model to reduce the phase unwrapping error. Guo \textit{et al.} \cite{guo2023unifying} built a deep learning model that can perform three types of TPU and lowered the phase unwrapping error by using deep learning. Suresh \textit{et al.} \cite{suresh2021pmenet} introduced a method of enhancing the phase map obtained from the Fourier transform through deep learning. Yan \textit{et al.} \cite{yan2020wrapped} designed a denoising network and conducted a study to remove the noise of the phase map.

Lastly, some efforts have been dedicated to deriving unwrapped phases directly from fringe images. Zhang \textit{et al.} \cite{zhang2022deep} designed DLALNet which predicts an unwrapped phase from 3 fringe images. Li \textit{et al.} \cite{li2022deep} and Nguyen \textit{et al.} \cite{nguyen2022accuratergb} introduced frameworks that could obtain an unwrapped phase map from a single fringe image with a composite fringe pattern or RGB pattern, respectively. Nguyen \textit{et al.} \cite{nguyen2023generalized} also suggested a framework to obtain an unwrapped phase from a single fringe image by predicting phase-shifted fringe images and a reference phase.

%% file: secs/3_background.tex
\section{Backgrounds} \label{sec:background}

\begin{figure}
    \includegraphics[width=\textwidth]{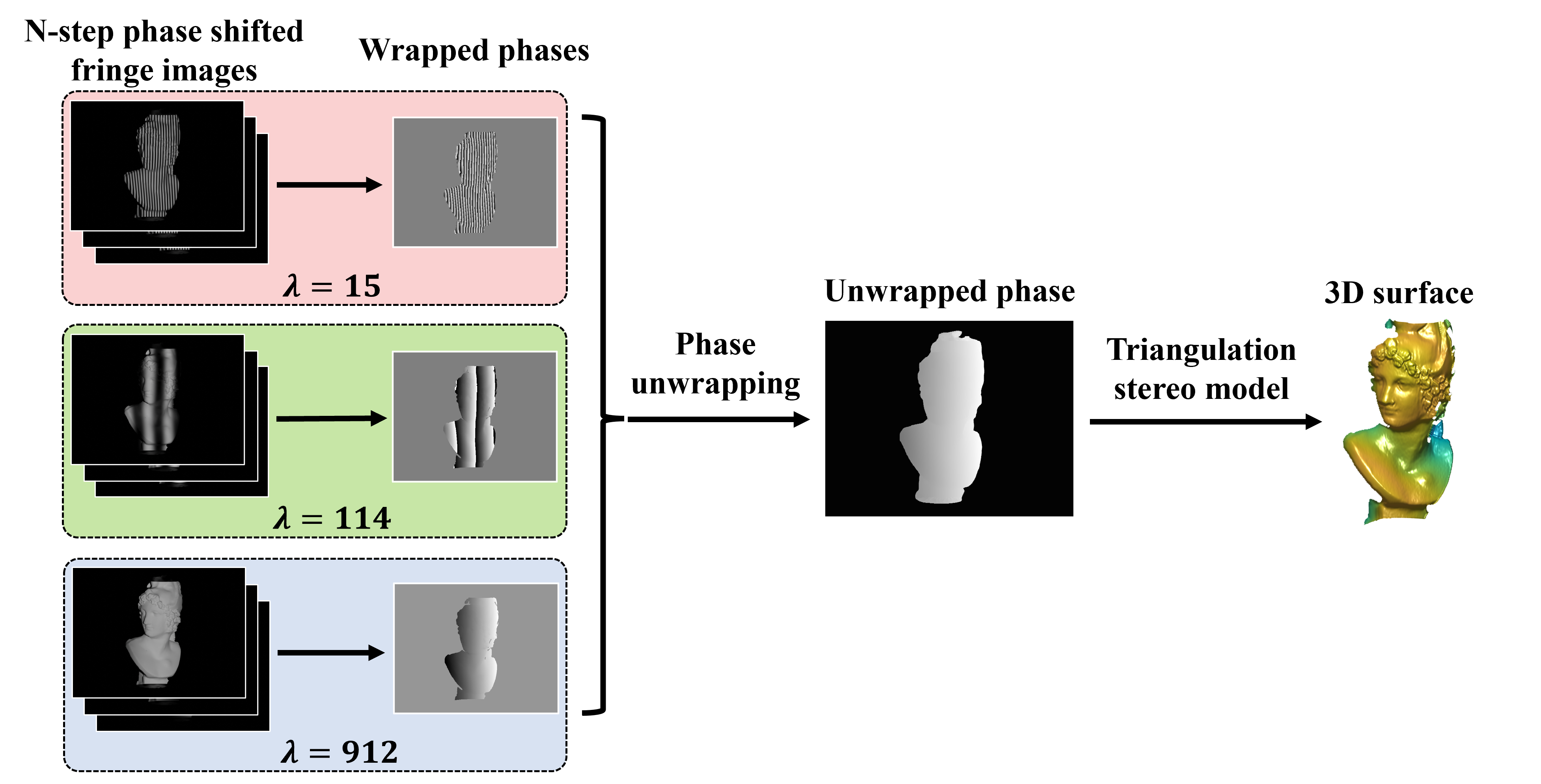}
    \caption{Framework of 3D reconstruction using the conventional FPP-based technique.}
    \label{fig:convenfpp}
\end{figure}

In this section, we introduce the foundational principles of the FPP system, which are essential for comprehending our proposed method. The 3D surface measurement system based on FPP fundamentally comprises a camera and a projector. During the capturing stage, the projector projects fringe patterns onto the target object, and the camera captures distorted fringe images. These distorted fringe images undergo fringe analysis to be transformed into the phase domain, ultimately facilitating the 3D surface reconstruction. Fig.~\ref{fig:convenfpp} depicts the workflow of the conventional FPP-based 3D shape reconstruction technique.

\subsection{Phase Shifting Algorithms}
\label{sec:PS algorithm}
A digital fringe projection (DFP) system uses a digital video projector to project computer-generated sinusoidal (fringe) patterns. The computer-generated fringe patterns can be described as
\begin{equation}
\label{equation:fringe1}
I(u, v) = 255/2[1+\text{cos}(2\pi i / \lambda + \delta)],
\end{equation}
where $I$ represents the intensity value, $(u,v)$ refers to the pixel index, $\lambda$ is a fringe pitch, the number of pixels per period, and $\delta$ is the amount of phase-shifting.
The N-step phase-shifting (PS) algorithm is the most widely used method for phase retrieval due to its accuracy and robustness to noise \cite{zuo2018phase}. Fringe images based on a phase-shifting algorithm can be described as  
\begin{equation}
    \label{equation:fringe2}
        \begin{split}
            I_n(u, v) &= I'(u,v) + I''(u,v)\text{cos}(\phi(u,v)+\delta_n), \\
            \delta_n &= \frac{2\pi (n-1)}{N}, ~n=1,2, \ldots ,N,
        \end{split}
\end{equation}
where $(u,v)$ represents the camera pixel coordinate, $I', I''$ represents the average intensity and intensity modulation, respectively, and $\phi$ is the phase. By solving Eq.~\ref{equation:fringe2} with least square algorithms, the wrapped phase from the N-step phase shifting algorithm can be calculated as
\begin{equation}
    \label{equation:nstep_wrapped_phase}
    \phi(u, v) = -\text{tan}^{-1} 
    \left[\frac{\sum_{n=0}^{N-1}{I_n(u, v) \text{sin}(2\pi n/N)}}{\sum_{n=0}^{N-1}I_n(u, v)\text{cos}(2\pi n/N)}\right] = -\text{tan}^{-1}\left[\frac{M(u,v)}{D(u,v)}\right],
\end{equation} 
where $M$ and $D$ represent the numerator term and the denominator term consisting of the wrapped phase.
The calculated phase values (so-called wrapped phase) are limited to a range of $(-\pi, \pi]$ due to the property of an arc-tangent function. To obtain an absolute phase, an additional process called \textit{phase unwrapping} is needed.

\subsection{Temporal Phase Unwrapping}
Since the wrapped phase values range from $-\pi$ to $\pi$, it is necessary to assign a fringe order to each repeating pattern to obtain continuous and unique phase values.
If the fringe order map is obtained from the phase unwrapping algorithm, the absolute phase map can be described as 
\begin{equation}
    \label{equation:phase_unwrapping}
    \Phi(u, v) = \phi(u, v) + 2\pi \times K(u, v),
\end{equation}
where $K$ represents the fringe order map.

There are some types of TPUs: Multi-frequency TPU (MF-TPU) \cite{huntley1993temporal}, Multi-wavelength TPU (MW-TPU) \cite{cheng1984two}, and Number Theoretical TPU \cite{gushov1991automatic}. The common idea of these methods is to unwrap the phase with the aid of one or more additional phases (reference phase) with different fringe periods. Among these methods, the MF-TPU method is widely used because of its unwrapping reliability.

If there are given two wrapped phases denoted as $\phi_h, and \phi_l$, its corresponding absolute phase maps should have the following relationship.
\begin{equation}
    \label{equation:prop2pitch}
    \Phi_h (u, v) \lambda_h = \Phi_l (u,v) \lambda_l.
\end{equation}
Here, subscripts $h$ and $l$ represent the high-frequency and low-frequency, respectively. The frequency of the fringe pattern is defined as $W/\lambda$, where $W$ is the width of the projector's resolution. 

For the case of two-frequency TPU, $\phi_l$ is retrieved by using a set of unit frequency patterns, so the one pitch of the fringe pattern covers the whole surface. Therefore, no phase unwrapping is required for $\phi_l$, and $\Phi_l = \phi_l$, so the low-frequency phase can be used as a reference phase without an additional process. By using Eq.~\ref{equation:phase_unwrapping} and Eq.~\ref{equation:prop2pitch}, the fringe order map $K_h(u,v)$ can be obtained by the following equation.
\begin{equation}
    \label{equation:mftpu}
    K_h(u, v) = \text{Round}\left[\frac{(\lambda_l / \lambda_h) \Phi_l(u,v) - \phi_h(u,v)} {2 \pi}\right].
\end{equation}

The following is the case where three or more frequencies are used. Again, the absolute phase is determined from the wrapped phase of the highest frequency, and the lower frequency wrapped phases are used as reference phases to obtain the fringe order map. Assuming that P phase maps are used, the phase unwrapping process can be described as
\begin{equation}
    \label{equation:mftpu_3f}
    K_i(u, v) = \text{Round}\left[\frac{(\lambda_{i-1} / \lambda_i) \Phi_{i-1}(u,v) - \phi_i(u,v)} {2 \pi}\right],
\end{equation}
where subscript $i$ represents the $i$th frequency with $i = {2, 3, ..., P}$. In this case, the frequency should satisfy $f_P > f_{P-1} > ... > f_{1}$ or $\lambda_1 > \lambda_2 > ... > \lambda_P$. Also, as in the case of two-frequency TPU, $\Phi_1 = \phi_1$ must be satisfied. The absolute phase is calculated in the recursive order.

\subsection{Minimum-phase method} \label{sec:minimum_phase}

 As seen in Eq.~\ref{equation:mftpu}, the noise in the reference phase is scaled up by a factor of $\lambda_l / \lambda_h$, so the phase unwrapping error amplifies as the value of the low-frequency fringe pitch is significantly large. One approach to addressing these issues is by utilizing geometric system constraints. Hyun suggested the enhanced method by using geometric constraints of the system \cite{hyun2016enhanced}. A synthetic phase map, denoted as $\Phi_{min}$, is derived from geometric constraints of the system at a given virtual depth plane at $z=z_{min}$, which is the closest depth of interest.
\begin{equation}
    \label{equation:zmin1}
    \Phi_{min} = f(z_{min}, \lambda, \theta),
\end{equation}
where $\theta$ represents the calibration parameters which contain the focal length, principle points, and position, rotation vectors of the camera and projector.

In Hyun's method, the low-frequency is not 1, so the low-frequency phase (reference phase) has phase ambiguities. $\Phi_{min}$ supports the reference phase of the z-min method ($\phi_z$) to remove the phase ambiguities. Low-frequency fringe order map $K_z$ can be determined as
\begin{equation}
    \label{equation:zmin2}
    K_z (u,v) = \text{Ceil}\left[\frac{\Phi_{min} - \phi_z}{2\pi}\right],
\end{equation}
where $\text{Ceil}[\cdot]$ function computes the smallest integer that is greater than or equal to a specified number. By adding the fringe order map multiples $2\pi$ to the low-frequency wrapped phase, the phase ambiguities of the low-frequency wrapped phase are removed, and the reliability of the absolute phase is enhanced compared to unit frequency. The rest of the process is the same as the conventional MF-TPU (Eq.~\ref{equation:mftpu}). By enhancing the low-frequency wrapped phase, the phase unwrapping error decreases. However, in the case of using geometric constraints, there is a depth range limitation \cite{an2016pixel}. The depth range is proportional to the pitch of the fringe pattern. Therefore, it is necessary to choose an appropriate fringe pitch by considering the depth range of the system. 

\subsection{Depth reconstruction}

Through the preceding steps, the unwrapped phase of the measured object can be obtained, and this phase is converted into the height of the measured object using calibration data. Depending on the calibration method, it can be broadly categorized into the phase-height model and the triangular stereo model \cite{feng2021calibration}. In the phase-height model, the surface's height is represented as a function of the phase, i.e., $h(u,v) = f(u, v, \Phi, \theta)$, where $\theta$ is calibration coefficients. Here, the height is measured as the relative position from a reference plane defined by $h(u, v) = 0$. Typically, the phase-height model is approximated and utilized as either a linear model \cite{tavares2007linear} or a polynomial model \cite{chen2016accurate}.

In the triangular stereo model, the phase does not directly represent the height. Instead, it serves as a linkage between the camera pixel and the projector pixel. This model can be viewed as an extension of the stereo vision approach. The camera and projector in the DFP system follow pinhole models \cite{zhang2000flexible}, and their intrinsic and extrinsic parameters are calculated through system calibration. Given the known positions of both the camera and projector pixels, the 3D coordinates of an object point can be determined by triangulation. Here, we chose the triangular stereo model for depth reconstruction \cite{zhang2006novel}.

%% file: secs/4_method.tex
\section{Proposed Method}\label{sec:proposed method}

\begin{figure}[t]
    \centering
    \begin{subfigure}{\textwidth}
        \includegraphics[width=\textwidth]{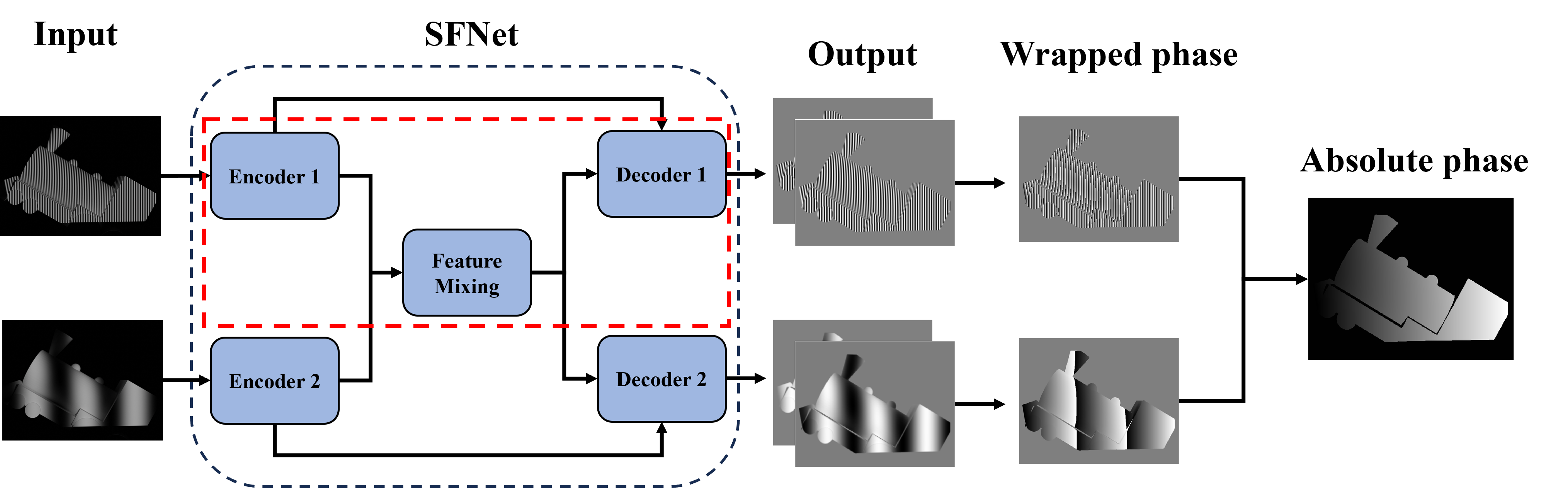}
        \caption{Pipeline of the proposed method}
        \label{fig:flowchart}
    \end{subfigure}
    \vfill
    \begin{subfigure}{\textwidth}
        \includegraphics[width=\textwidth]{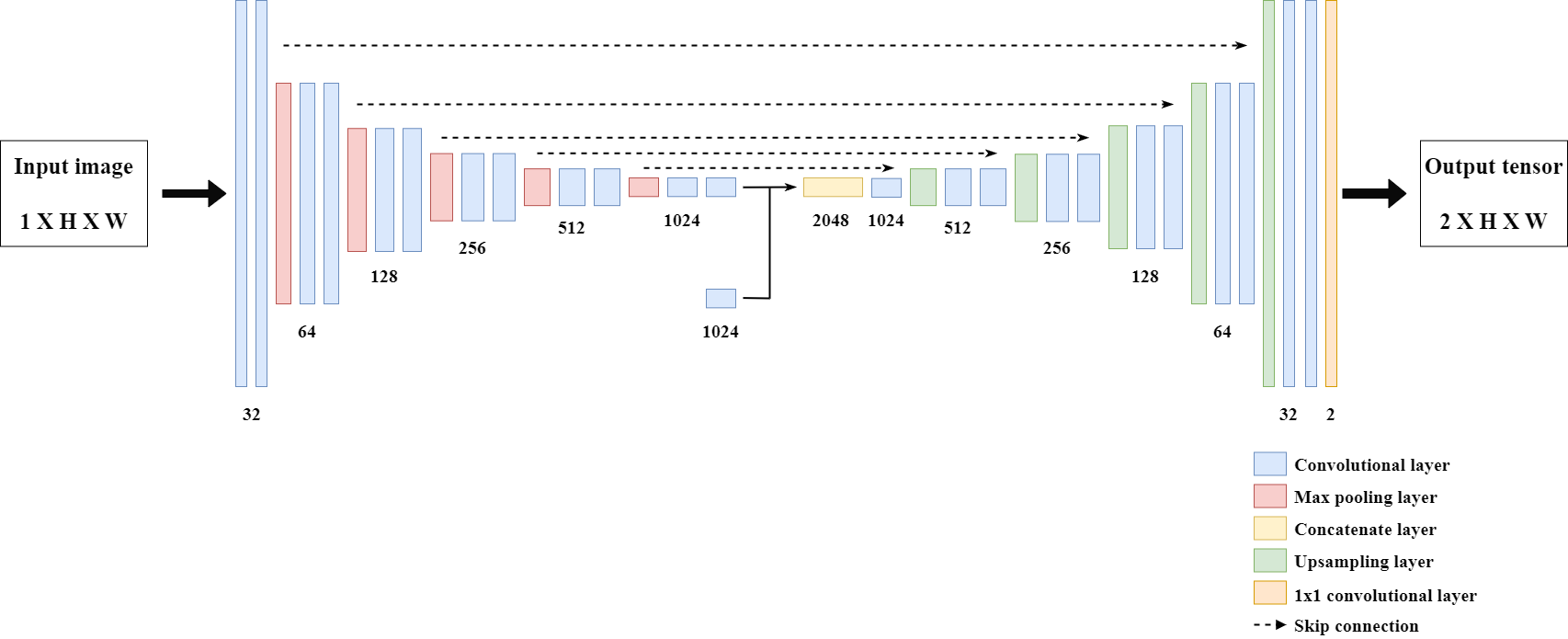}
        \caption{Architecture of SFNet}
        \label{fig:architecture}
    \end{subfigure}
    \caption{\textbf{Illustration of the proposed method.} The neural network structure of the red-dashed box (a) is schematized in (b).}
    \label{fig:proposedmethod}
\end{figure}

The traditional TPU method requires a set of fringe images to derive an absolute phase, which consequently slows down the measurement speed. Recognizing the inefficiencies associated with this, we introduce a novel framework based on deep learning. Our approach not only aims to streamline the process but also anticipates notable enhancements in accuracy, efficiency, and computational speed.

\subsection{Network Architecture}
\label{sec:architecture}
Fig.~\ref{fig:proposedmethod} shows the workflow of our research. To predict the absolute phase, our method employs two fringe images. Symmetric Fusion Network (SFNet), which is our proposed network, features two encoders and two decoders. Each encoder extracts a feature map from the corresponding fringe image. These feature maps are merged and each decoder predicts the corresponding wrapped phase from the merged feature map. The decoder does not predict the wrapped phase directly, but the numerator and denominator term of the wrapped phase instead. It is known that predicting a wrapped phase by dividing into the numerator and denominator term is more beneficial for achieving accurate phase retrieval than directly predicting the wrapped phase \cite{feng2019fringe, dou2022deep}.

The detailed architecture of SFNet is described in Figure~\ref{fig:architecture}. Structurally, SFNet is composed of two encoders and two decoders. The encoder-decoder structure is symmetrically designed. The encoder-decoder pair structure is similar to the traditional U-Net architecture \cite{unet}. Each encoder, upon receiving a fringe image, extracts features through 12 convolutional layers. Subsequent max-pooling layers halve the feature map's resolution but double its channel depth. After the encoder part,  a concatenate layer merges the feature maps from two encoders. Then the decoders take the merged feature map as an input. The decoder produces a wrapped phase corresponding to the input image. The model's output consists of the numerator and denominator of the wrapped phase, and the wrapped phase can be calculated through the arc-tangent function in Eq.~\ref{equation:nstep_wrapped_phase}.

\subsection{Refined reference phase}
\label{sec:refined_ref_phase}

\begin{figure}[hbt!]
    \centering
    \begin{subfigure}{\textwidth}
        \includegraphics[width=\textwidth]{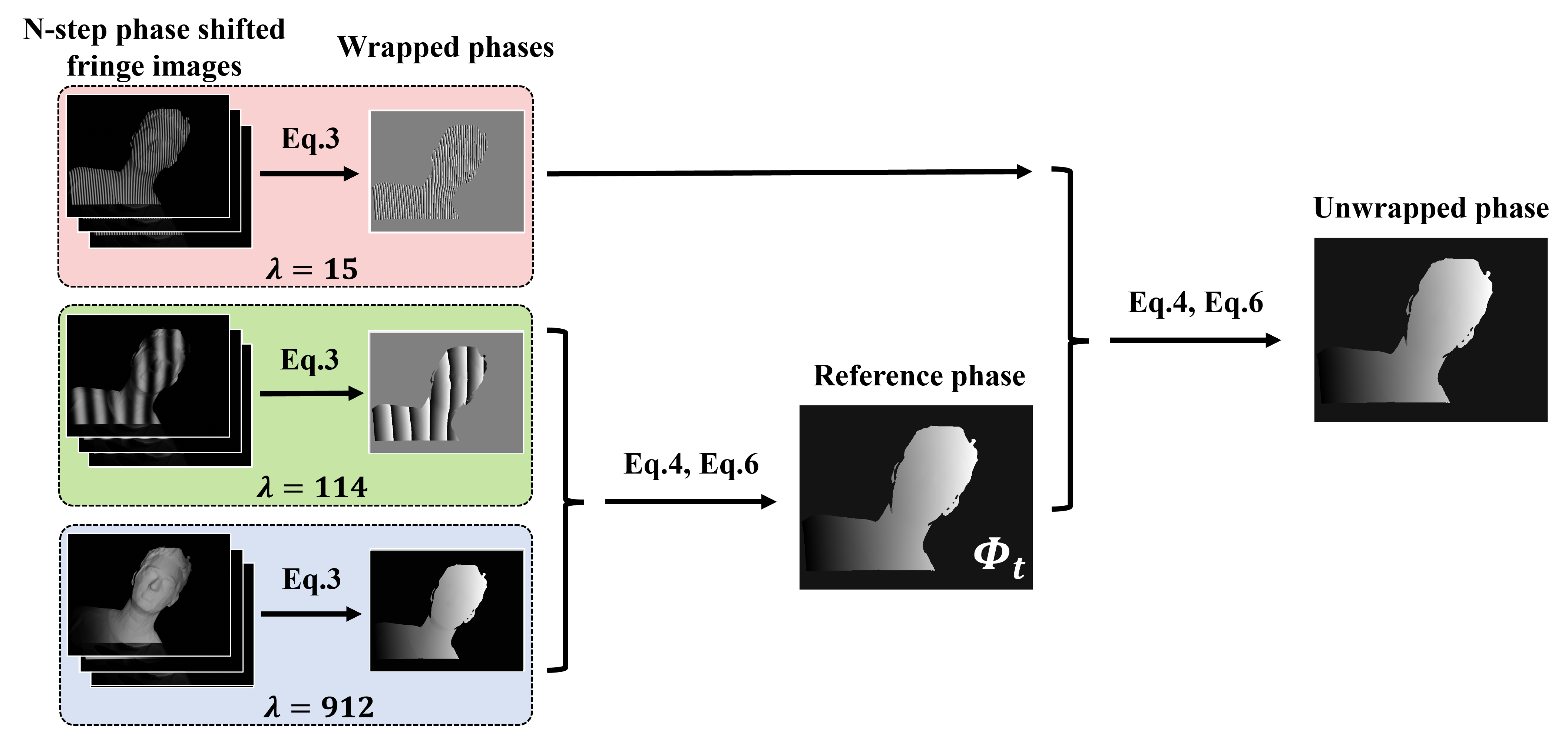}
        \caption{MF-TPU with three frequencies}
        \label{fig:3ftpumethod}
    \end{subfigure}
    \hfill
    \begin{subfigure}{\textwidth}
        \includegraphics[width=\textwidth]{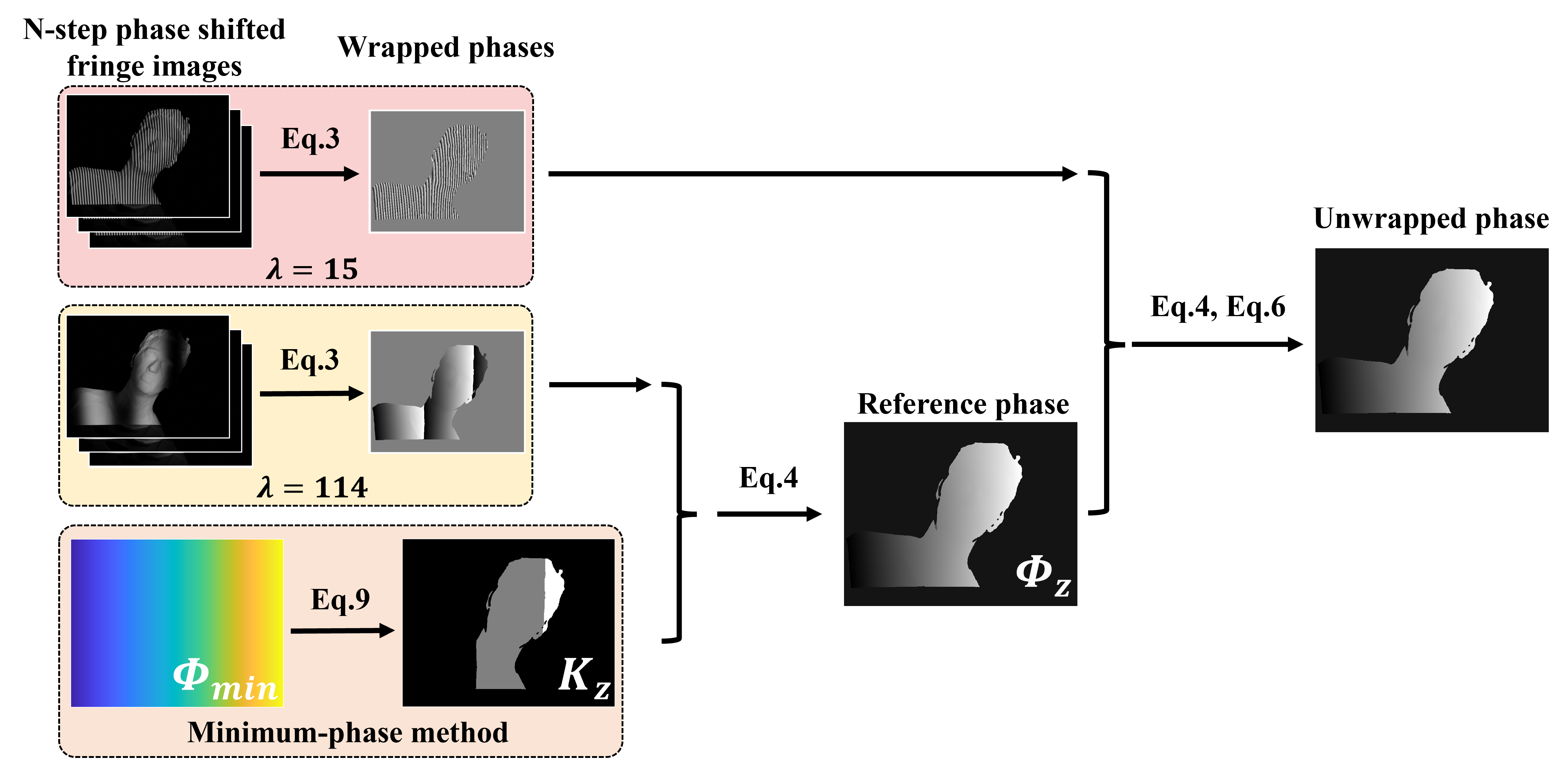}
        \caption{Modified MF-TPU with z-min method}
        \label{fig:zminmethod}
    \end{subfigure}
        \caption{\textbf{Frameworks of MF-TPU method and modified MF-TPU.} To implement the TPU method, a reference phase without phase ambiguity is essential. (a) In the conventional MF-TPU, a unit frequency fringe pattern is employed for this purpose. (b) The modified approach constructs a minimum phase using geometric constraints, thereby eliminating phase ambiguity.}
        \label{fig:tpuvszmin}
\end{figure}

In principle, at least two wrapped phases are required for TPU methods, but to minimize the phase unwrapping errors, our research uses three wrapped phases, $\lambda = \{15, 114, 912 \}$, to establish robust ground truth. However, SFNet's design limits it to predicting just two wrapped phases, which could potentially compromise phase reliability. To obtain a high-quality absolute phase map by predicting only two phase maps, it is important to increase the quality of the reference phase. We increased the quality of the reference phase using the following two steps.

\begin{figure}[hbt!]
    \centering
    \includegraphics[width=\textwidth]{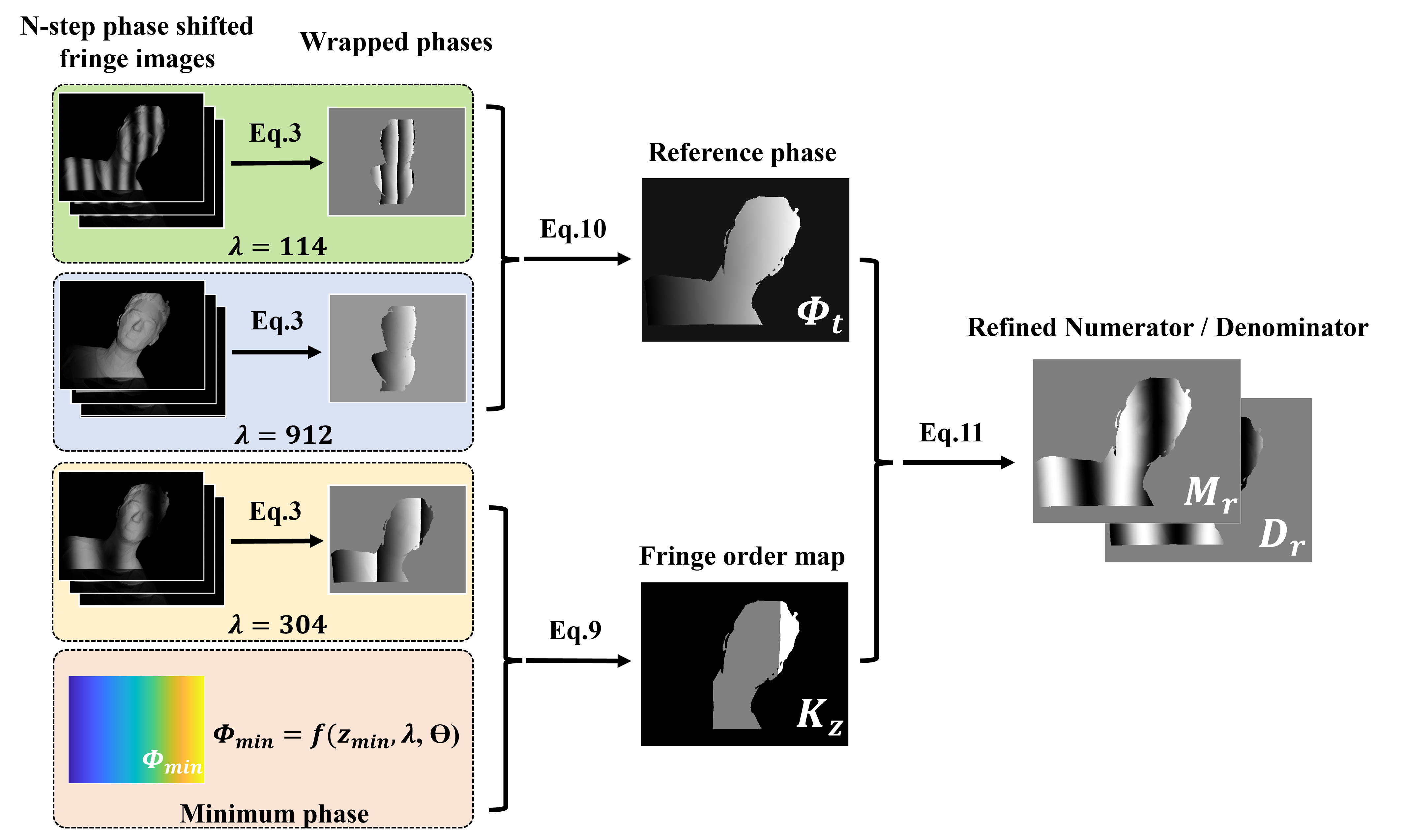}
    \caption{\textbf{Framework of obtaining the refined reference phase.}}
    \label{fig:ppmethod}
\end{figure}

\paragraph{Step 1: Minimum-phase (z-min) method.} In this context, we employ a modified TPU incorporating geometric constraints, which is described in Section~\ref{sec:minimum_phase}. Fig.~\ref{fig:tpuvszmin} explains the procedures of the conventional MF-TPU and the introduced method, and Fig.~\ref{fig:ppmethod} depicts the procedure for acquiring the refined reference phase. 

In the proposed method, additional images with frequencies not used in the three-frequency TPU are employed to obtain a reference phase of higher quality. The reason for using additional images is that the minimum-phase method has a depth range inversely proportional to the fringe pitch. In the conventional process, to apply geometric constraints, a synthetic phase with a fringe frequency of 8 needs to be generated. However, this is challenging for dynamic scenes due to the very limited allowable depth range. To ensure an adequate depth range, we use additional fringe images with a frequency of 3 $(\lambda = 304)$. We captured phase-shifted fringe images with a frequency of 3, and based on these, obtained the corresponding fringe order map $K_z$ using the wrapped phase and Eq.~\ref{equation:zmin2}.

\begin{figure}[hbt!]
    \centering
    \begin{subfigure}{\textwidth}
        \includegraphics[width=\textwidth]{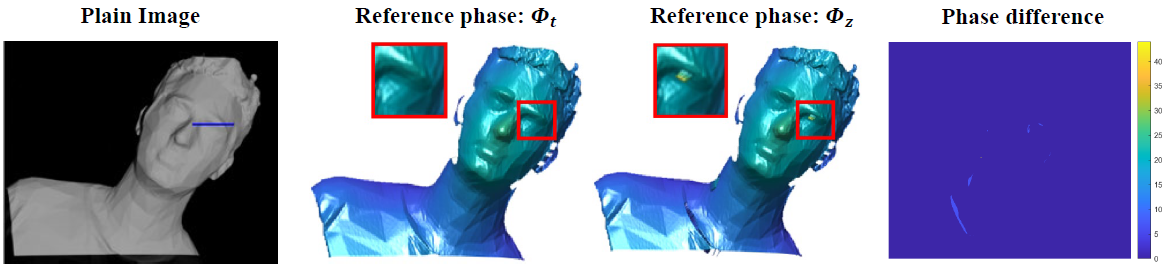}
        \caption{}
        \label{fig:refined1}
    \end{subfigure}
    \hfill
    \begin{subfigure}{\textwidth}
        \includegraphics[width=\textwidth]{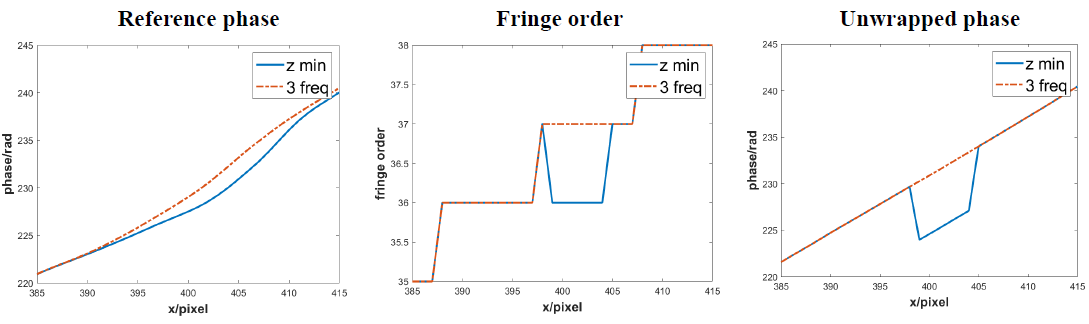}
        \caption{}
        \label{fig:refined2}
    \end{subfigure}
        \caption{\textbf{Comparison of the two reference phases obtained in Fig.~\ref{fig:tpuvszmin}.} (a) 3D plot and phase difference between the three-frequency TPU method shown in Fig.~\ref{fig:3ftpumethod}) and minimum-phase method shown in Fig.~\ref{fig:zminmethod}. (b) Cross-lines of the reference phase, fringe order, and unwrapped phase are marked in the plain image in (a).}
        \label{fig:refinedph}
\end{figure}

\paragraph{Step 2: Refined wrapped phase.} The frequency of the reference phase was set to 3 to utilize the minimum-phase method in a wider depth range. However, in the process of building the ground truth, the used frequencies are 64 $(\lambda = 15)$, 8 $(\lambda = 114)$, and 1 $(\lambda = 912)$. Since the phase map of unit frequency is used to eliminate the phase ambiguities of the phase map with frequency 8, the frequency of the reference phase used in three-frequency TPU is 8. It is widely known that the higher the frequency of the fringe pattern, which means the thinner the fringe pitch, the more details can be expressed in the phase information. Therefore, the reference phase in the process of the three-frequency TPU has a higher quality than the reference phase of the minimum-phase method. As shown in Fig.~\ref{fig:refinedph}, there are a few spike errors or phase errors in the minimum-phase method. To reconstruct more details of the target object, we refined the reference phase of the minimum-phase method.
The procedure described above is as follows. Here, the subscripts 1 and 8 represent the corresponding frequency. First, using the minimum-phase method, a fringe order map corresponding to the reference phase of z-min method $(K_z)$ can be calculated by Eq.~\ref{equation:zmin1}. Also, the reference phase used in three-frequency TPU is obtained as follows.
\begin{equation}
    \label{equation:3freq_mid}
    \begin{split}
        \Phi_1 &= \phi_1, \\ \Phi_8 = \phi_8 + 2\pi \times & \text{Round}(\frac{(\lambda_1 / \lambda_8) \Phi_1 - \phi_8}{2\pi}).
    \end{split}
\end{equation}
Here, we denote $\Phi_8$ as $\Phi_t$, the meaning of the reference phase of the three-frequency TPU method. To enhance the reference phase, we scaled $\Phi_t$ with a range of $\Phi_z$ and subtracted $K_z$ obtained earlier to get a wrapped phase, $\phi_r$ with more detail than before. By decomposing the refined wrapped phase obtained in this way by sine and cosine terms again, the numerator and denominator terms of higher quality than before can be obtained as,
\begin{equation}
    \label{equation:refined_nd}
        \begin{split}
            \Phi_{r} = \frac{\lambda_8}{\lambda_3} \times \Phi_t, ~\phi_{r} = \Phi_{r} - 2\pi \times K_z,   \\ M_r = -\text{sin}(\phi_{r}), D_r = \text{cos}(\phi_{r}).
        \end{split}
\end{equation}
If we denote a wrapped phase of $\lambda = 15$ as $\phi_h$, the corresponding unwrapped phase can be obtained through Eq.~\ref{equation:mftpu} and Eq.~\ref{equation:zmin1}, using only two wrapped phases, $\phi_h$ and $\phi_r$. This results in a quality equivalent to the TPU method using three frequencies. As a result, our approach aims to predict $M_h, D_h, M_r, D_r$, which form $\phi_h$ and $\phi_r$, through a neural network from two fringe images.

\subsection{Training}
\label{sec:loss}
The loss function of SFNet consists of three types of losses: 1) phase loss, 2) phase consistency loss, and 3) geometric constraint loss.
 
\paragraph{Phase Loss} The phase loss represents the loss between the output from decoders and the corresponding ground truth. Also, the phase loss includes the unwrapped phase error between the prediction and the ground truth. The phase loss can be described as
\begin{equation}
    \label{equation:phaseloss}
    \mathcal{L}_{\text{phase}} = (\tilde{M_h} - M_h)^2 + (\tilde{D_h} - D_h)^2 + (\tilde{M_l} - M_l)^2 + (\tilde{D_l} - D_l)^2 + (\tilde{\Phi} - \Phi)^2,
\end{equation}
where $\tilde{M}, \tilde{D}, \tilde{\Phi}$ represents the predicted numerator and denominator of the wrapped phase, and unwrapped phase, respectively.

\paragraph{Phase Consistency Loss} Phase consistency loss constraints the high frequency wrapped phase to have the same gradient as the unwrapped phase. As mentioned before, the unwrapped phase is calculated by adding integer value (fringe order) multiples $2\pi$ to the wrapped phase, so the fringe order map can not affect the gradient of the unwrapped phase. So, it is obvious that the gradient of the wrapped phase and unwrapped phase are the same. We calculated the gradient map of the predicted high-frequency wrapped phase and compared the loss with the corresponding ground truth. The phase consistency loss can be described as
\begin{equation}
    \label{equation:phaseconsloss}
    \mathcal{L}_{\text{consist}}= (\frac{\partial \tilde{\phi}}{\partial u} - \frac{\partial \Phi}{\partial u})^2 + (\frac{\partial \tilde{\phi}}{\partial v} - \frac{\partial \Phi}{\partial v})^2,
\end{equation}
where $\tilde{\phi}$ represents the predicted wrapped phase, and $\partial$ is the partial derivative along the camera pixel coordinate $(u, v)$.

\paragraph{Geometric Constraint Loss} Geometric constraint loss is related to the spike noise of the recovered 3D surface. Our network is a kind of regression model, and regression methods typically incorporate global optimization techniques, which may result in smoothing effects of the phase map. The small error in the phase map can induce the phase unwrapping error. To alleviate the phase unwrapping error, we constrain the pixel value to its adjacent pixel value, to reduce the spike error. The geometric loss can be described as
\begin{equation}
    \label{equation:geoconstloss}
    \mathcal{L}_{\text{geo}}= \sum(\Phi(u_c, v_c) - \Phi(u_c + 1, v_c))^2 + \sum(\Phi(u_c, v_c) - \Phi(u_c, v_c + 1))^2.
\end{equation}

The total loss function of SFNet is defined as

\begin{equation}
    \label{equation:totalloss}
    \mathcal{L}_{\text{total}}=\lambda_{\text{phase}} \mathcal{L}_{\text{phase}} + \lambda_{\text{consist}} \mathcal{L}_{\text{consist}} + \lambda_{\text{geo}}\mathcal{L}_{\text{geo}},
\end{equation}
where the $\lambda_{\text{phase}}, \lambda_{\text{consist}}$ and $\lambda_{\text{geo}}$ are the weights of each loss function. We set $ \lambda_{\text{phase}}=1, \lambda_{\text{consist}}=1e-2, \lambda_{\text{geo}}=1e-6$ with the experimental results.

%% file: secs/5_dataset.tex
\section{SynthFringe Dataset} \label{dataset}

\begin{figure}[hbt!]
    \centering
    \begin{subfigure}[b]{0.45\textwidth}
        \includegraphics[width=\textwidth]{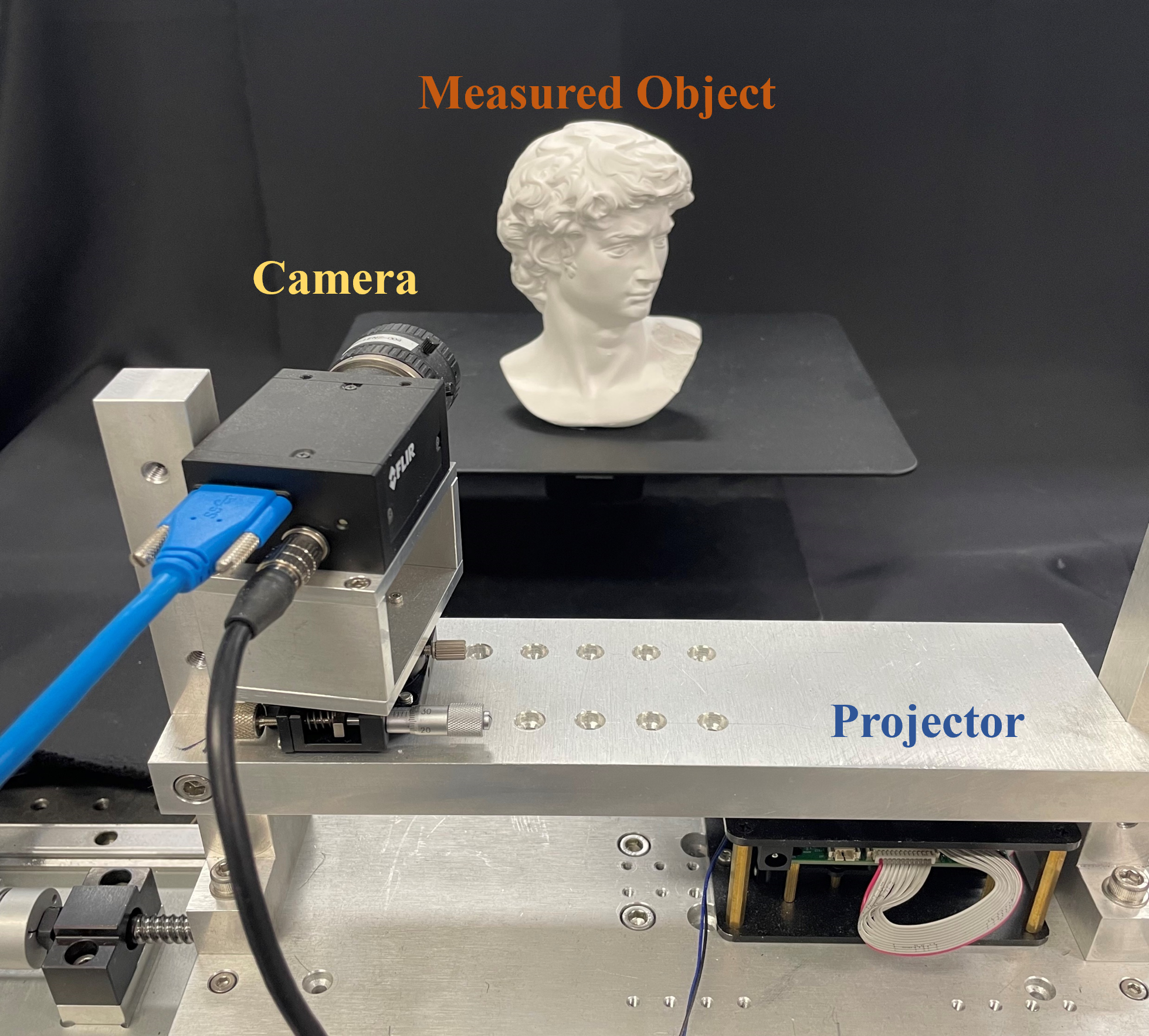}
        \caption{}
        \label{fig:virtualfpp1}
    \end{subfigure}
    \hfill
    \begin{subfigure}[b]{0.45\textwidth}
        \includegraphics[width=\textwidth]{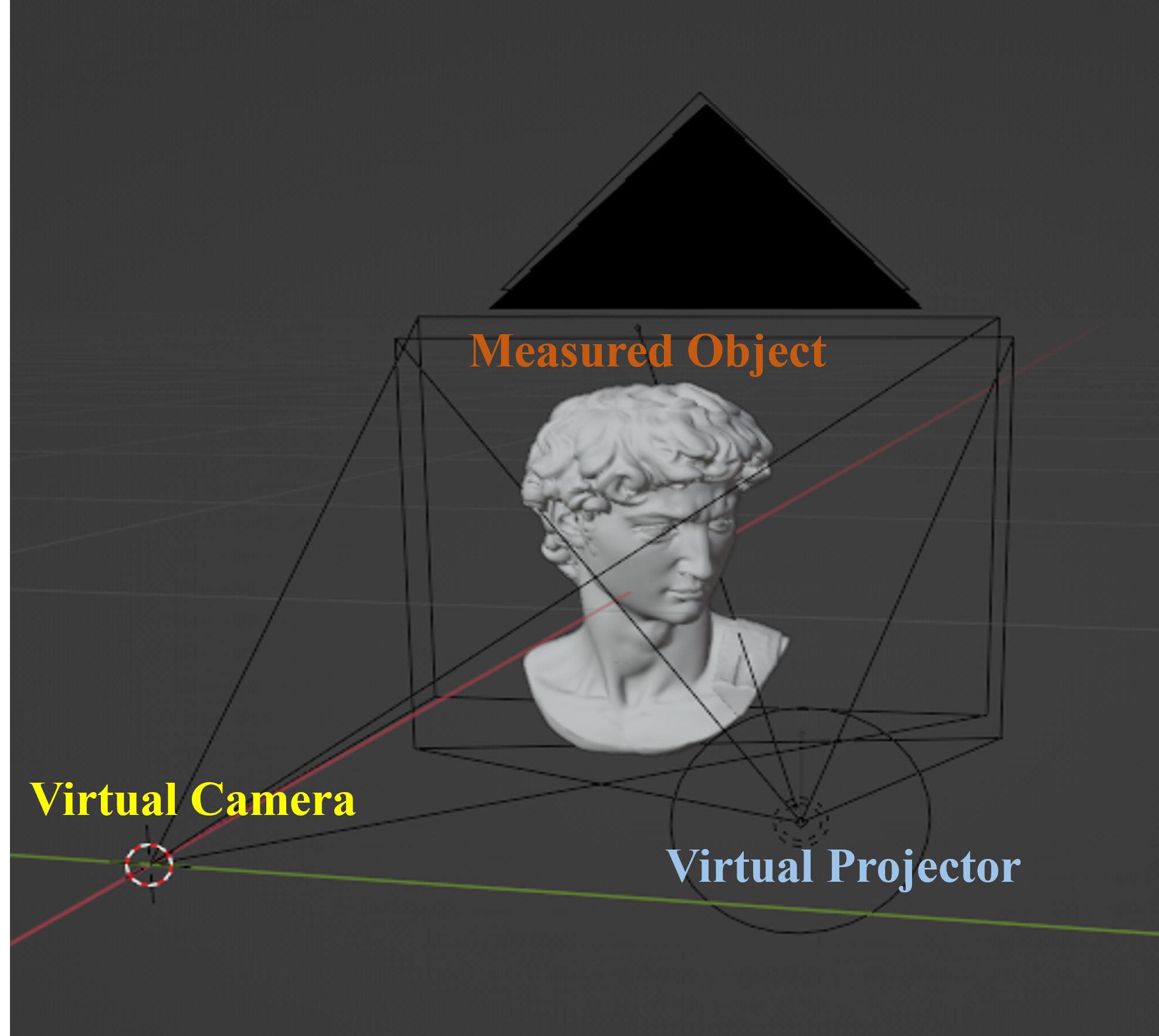}
        \caption{}
        \label{fig:virtualfpp2}
    \end{subfigure}
    \caption{\textbf{Illustration of FPP system.} (a) The setting of the FPP system in the real world. (b) Virtual FPP system by using CG software}
    \label{fig:virtualFPP}
\end{figure}

\begin{figure}[hbt!]
    \includegraphics[width=\textwidth]{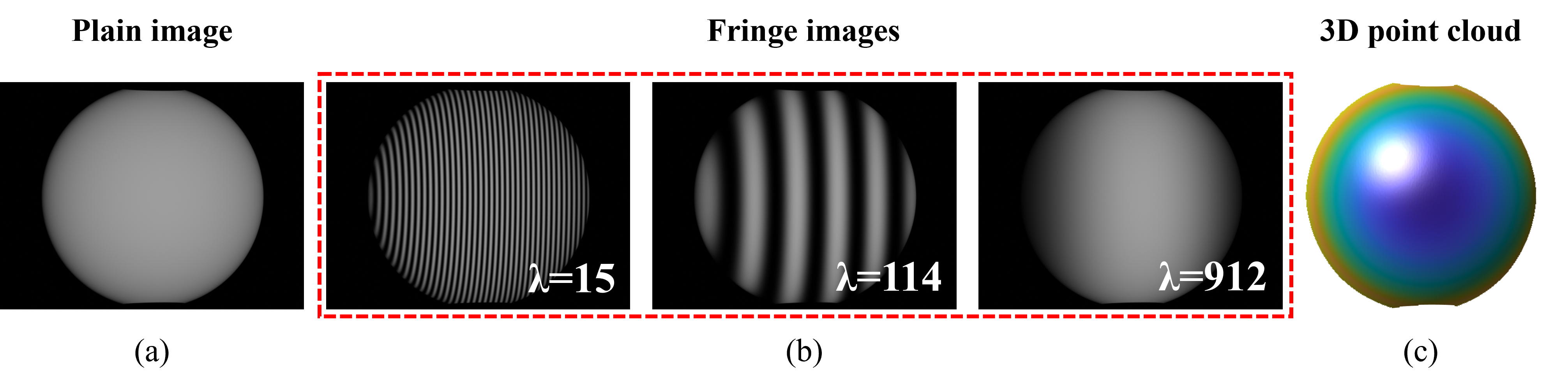}
    \caption{To validate the synthetic environment, we captured images of an ideal sphere. (a) The plain image of a sphere with a known dimension (radius = 100 mm). (b) Corresponding fringe images of the sphere with $\lambda=\{15,114,912\}$, respectively. (c) Reconstructed 3D point cloud of the sphere, where the calculated radius is $100.3970~mm$}
    \label{fig:spherefitting}
\end{figure}


For deep learning, it is important to build up a large dataset. Our framework utilizes a total of four frequencies $\lambda=\{15,114,304,912\}$ for training. Since datasets with such diverse frequencies were not previously available in the public domain, we had to construct our dataset from scratch. Table~\ref{table:comparison_dataset} compares the existing frameworks of the FPP system using deep learning that dataset has been released. 

\begin{table}
    \centering
    \begin{tabular}{c | c c c}
        \hline
        Datasets & Image pairs & Fringe image & Other content \\
        \hline \hline
        R. Zhang \cite{zhang2022deep} & 800 & \checkmark & WP ($\mathit{M}$, $\mathit{D}$) \\
        H. Nguyen \cite{yu2022untrained} & 1,025 &  & WP + FOM \\
        S. Bai \cite{bai2022deep} & 16,884 &  & WP + FOM \\
        H. Nguyen \cite{nguyen2023learning} & 1,500 & \checkmark & FOM \\
        H. Nguyen \cite{nguyen2023generalized} & 2,048 & \checkmark & UPH \\
        \hline
        SynthFringe (Ours) & 18,000 & \checkmark & WP \\
        \hline
    \end{tabular}
    \captionof{table}{\textbf{Comparison of Fringe Pattern Datasets}. WP: Wrapped phase, $\mathit{M}$, $\mathit{D}$: Numerator and Denominator term of wrapped phase (Eq.~\ref{equation:nstep_wrapped_phase}), FOM: Fringe order map, and  UPH: Unwrapped phase.}
    \label{table:comparison_dataset}
\end{table}

Our newly constructed dataset, named the ``SynthFringe'' dataset, not only contains a greater number of image pairs compared to other datasets but also encompasses images of diverse frequencies. Moreover, we made efforts to design our dataset with a variety of scenes to make it applicable to more general situations. For instance, while the dataset collected by S. Bai et al. \cite{bai2022deep} is extensive, it predominantly focuses on dental models, making it less suitable for general research purposes. In this section, we discuss the composition of the dataset used in the proposed method and the validity of the data. 

\begin{figure}
    \centering
    \begin{subfigure}{0.75\textwidth}
        \includegraphics[width=\textwidth]{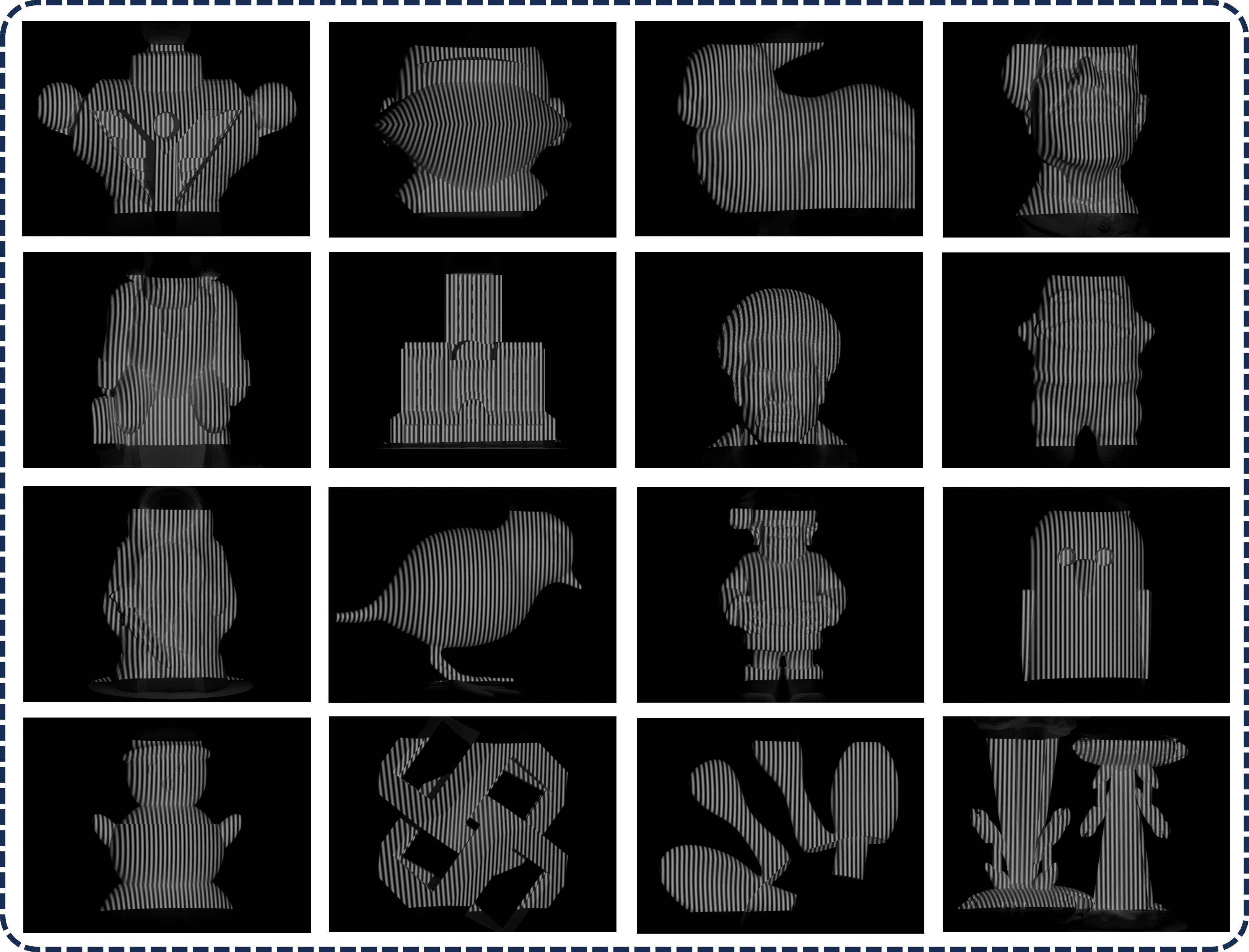}
        \caption{}
        \label{fig:datavis1}
    \end{subfigure}
    \hfill
    \begin{subfigure}{0.75\textwidth}
        \includegraphics[width=\textwidth]{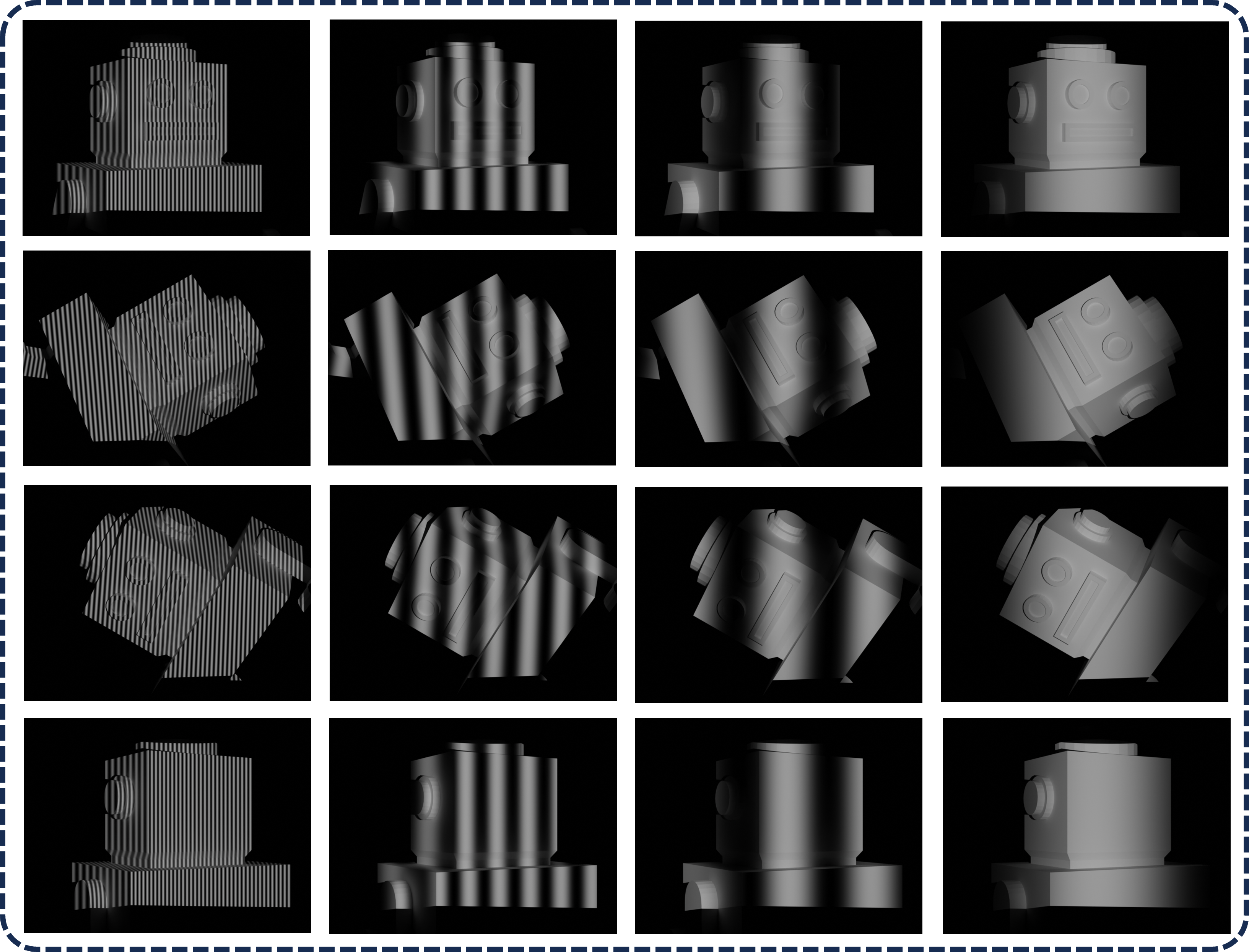}
        \caption{}
        \label{fig:datavis2}
    \end{subfigure}
    \caption{\textbf{Examples of ``SynthFringe'' dataset.} (a) Representative samples in the dataset. (b) Examples of the collected data, where each column represents the same fringe pattern and each row represents the same view}
    \label{fig:exampledataset}
\end{figure}

\subsection{Synthetic Data Generation}

\begin{table}[hbt!]
    \begin{center}
        \begin{tabular}{c | c c}
             \hline
                Parameters & Camera & Projector \\
            \hline
                Position $(m)$ & $(0, 0, 0)$ & $(-0.15, 0.045, 0)$\\
                Rotation $(^{\circ})$ & $(90, 0, 90)$ & $(90, 0, 95)$ \\
                Resolution $(pixel)$ & $640\times480$ & $912\times1140$ \\
            \hline
        \end{tabular}
    \caption{\textbf{Parameters of Virtual FPP System}}
    \label{table:paramsFPP}
    \end{center}
\end{table}

We build a virtual system using computer graphics (CG) software, Blender. Many other studies have already established an FPP system through a virtual environment \cite{martinez2022environment, zheng2020fringe, puljvcan2022simulation, quintero2022modeling}. Fig.~\ref{fig:virtualFPP} illustrates the implementation of an FPP system in Blender, simulating its functionality in the real world. By using a simulation platform, we could simulate the imaging processes more easily and cheaper than capturing images in the real world. The FPP system consists of a camera and a projector. We established our digital twin by using a blender's built-in camera module and a projector add-on. The resolution of the virtual camera and the projector are $640\times480$ and $912\times1140$, respectively. The focal length of the camera is $12~mm$. The other parameters of our system are shown in Table~\ref{table:paramsFPP}.

To verify our digital twin, we scanned the sphere with a known dimension (radius = $100~mm$). We computed the phase map by fringe projection and reconstructed the sphere by using the calibration data. Fig.~\ref{fig:spherefitting} displays images of a captured sphere for validation purposes, along with the reconstructed sphere. The calculated radius of the reconstructed sphere was $100.3970~mm$, with an error rate of $0.397~\%$. So it is proved that the virtual system can replace the real-world system. 

\subsection{Dataset Generation Pipeline}
\begin{figure}[t]
    \centering
    \begin{subfigure}{\textwidth}
        \includegraphics[width=\textwidth]{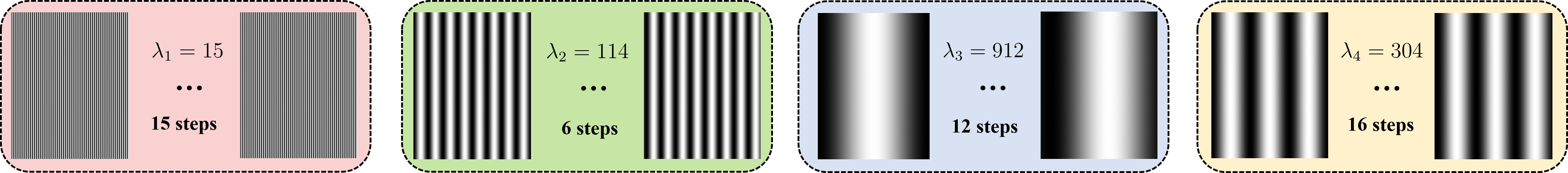}
        \caption{}
        \label{fig:dataset1_1}
    \end{subfigure}
    \hfill
    \begin{subfigure}{\textwidth}
        \includegraphics[width=\textwidth]{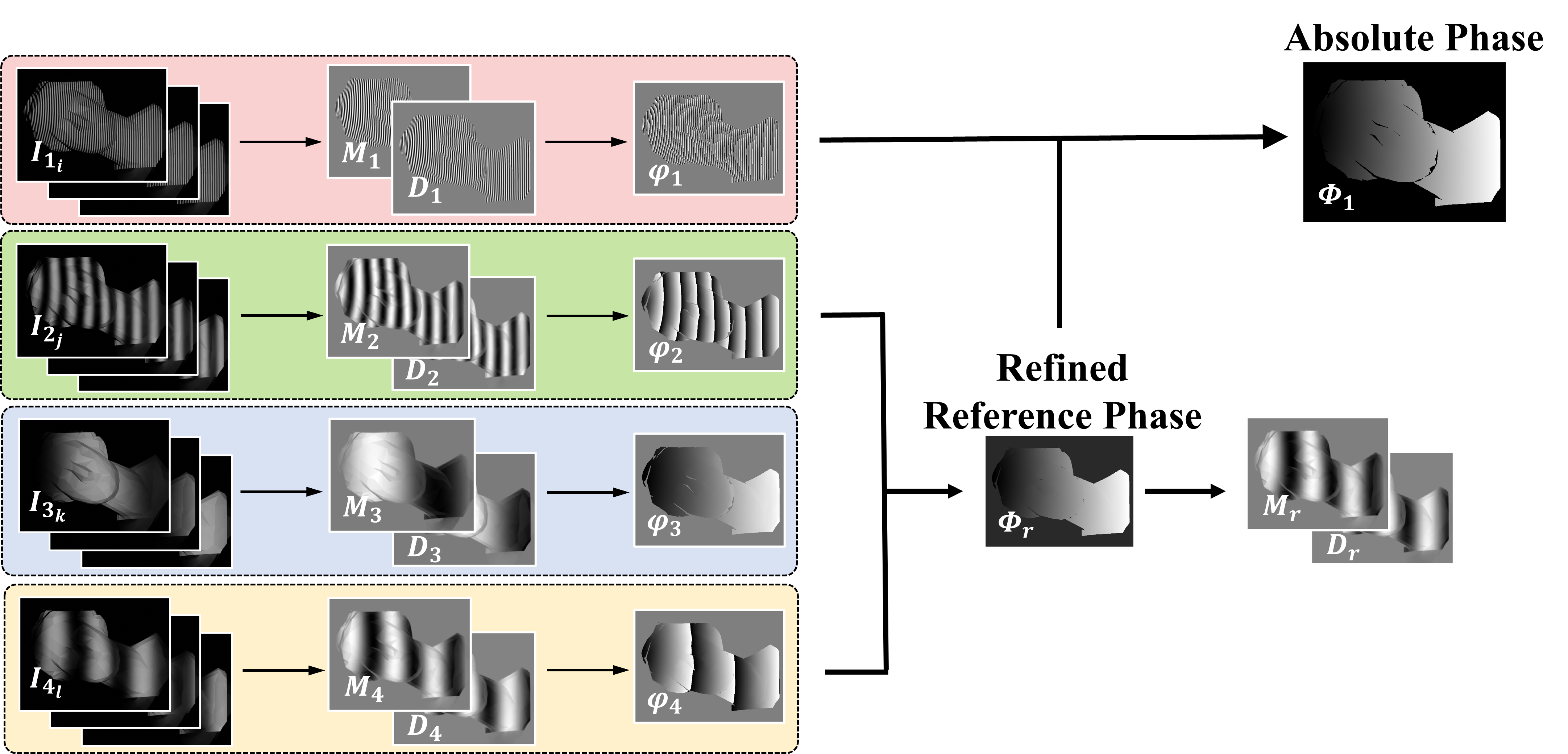}
        \caption{}
        \label{fig:dataset1_2}
    \end{subfigure}
    \caption{\textbf{Dataset generation procedure.} (a) Fringe patterns used for capturing data. (b) Process of configuring ground truth.}
    \label{fig:datasetprepare}
\end{figure}

We collected 500 CAD models from an open-source dataset, Thingi10K \cite{zhou2016thingi10k}. The selected objects are imported into the virtual system and positioned properly within the camera's field of view. To augment the dataset, we rotate each CAD model around the y-axis and z-axis, each for six times $(60^{\circ})$. So, as shown in Fig \ref{fig:datavis2}, we could get 36 different perspective views of each CAD model, producing 18,000 scenes. We split the dataset obtained from the virtual FPP system into 9:1 as a training dataset and validation dataset, respectively.

Fig.~\ref{fig:datasetprepare} above describes how to construct our dataset. Three frequencies, $\lambda = \{15, 114, 912\}$, are used to obtain a corresponding absolute phase. The absolute phase is calculated using Eq.~\ref{equation:phase_unwrapping} and Eq.~\ref{equation:mftpu_3f}. In addition, a phase map with $\lambda=304$ used in the minimum phase method is also obtained and used for training, which is described in Sec. \ref{sec:refined_ref_phase}. 

%% file: secs/6_experiments.tex
\section{Experiments} \label{sec:experiments}

\subsection{Implementation Details} 
As shown in Fig.~\ref{fig:architecture}, the architecture of SFNet comprises both encoder and decoder paths. The encoder path is structured with convolutional and max-pooling layers. Each convolutional layer possesses a kernel size of 3 and a stride size of 1. Moreover, zero padding is incorporated to maintain spatial dimensions. Following each convolutional layer, the ReLU activation function is applied. After traversing two convolutional layers for a single feature map, the max-pooling layer halves the feature map's resolution, while the filter depth of the convolutional layer doubles.
The two feature maps generated from each encoder are combined channel-wise via a concatenate layer. To prevent overfitting, a dropout function is added after the concatenate layer. The fused feature map then enters two decoders. The decoder path consists of convolutional and upsampling layers. Like the encoder path, the convolutional layers have a kernel size of 3, a stride size of 1, and zero-padding. The upsampling layer enlarges the feature map's resolution by a factor of two using transposed convolutional layers. At this stage, feature maps upsampled in the encoder are concatenated with those of identical resolution to utilize both low-level and high-level features simultaneously. The final step in the decoder path aligns the output channel (M, D) dimensions using a 1 by 1 convolutional layer.
 
We trained the model on a workstation with an AMD Ryzen Processor, 32GB RAM, and Nvidia RTX 4070Ti graphics card. The training of our model was implemented in Pytorch.
For training, the input fringe image is normalized to the range of [0, 1] by using the maximum and minimum values of the dataset.
The training configuration is as follows. The batch size is 8 and the maximum epoch is 250. The optimizer we choose is AdamW. Also, we reduced the learning rate using the scheduler class \textit{ReduceOnPlateau} in Pytorch.

In order to focus on the reliable points, we remove the invalid points by masking the data. The mask was derived from the intensity modulation of high-frequency fringe images. The Eq.~\ref{equation:masking} is the process of obtaining a mask.
\begin{equation}
\label{equation:masking}
\begin{split}
            I''_h(u, v) = \frac{2}{N_h}\sqrt{M_h(u,v)^2 + D_h(u,v)^2}, 
            \\ mask(u, v) = \left\{\begin{matrix} 1, ~I''_h(u,v)> Thrs \\ 0, ~I''_h(u,v)< Thrs
    \end{matrix}\right. ,       
\end{split}
\end{equation}
where $I''_h$ represents the intensity modulation of high-frequency fringe images, $N_h$ is the number of phase shifting (in our case, $N_h = 15$), and $Thrs$ refers to a threshold value to determine the fringe quality. The threshold value should be applied differently to each scene, but we set the $Thrs=8$ which is suitable for most of our dataset.

\subsection{Baselines}
To evaluate our framework we compare against other techniques for phase unwrapping, detailed below. All methods use fringe images as input and predict a corresponding unwrapped phase, but there is a difference in the number of fringe images. We implemented models using 1 to 3 fringe images, respectively, and trained them with our dataset.

SIDO \cite{nguyen2023generalized}, which is an abbreviation of single input-double output network, is a framework that predicts an unwrapped phase from a single image. The overall structure is similar to a typical U-Net, but the feature map from the encoder flows to two decoders, and each decoder predicts a high-frequency wrapped phase and an absolute phase, respectively. In SIDO, a high-frequency wrapped phase is obtained using predicted 4-step phase-shifted images. In this study, we modified the framework by changing from predicting 4-step fringe images to 5-step fringe images ($\lambda=15$). The 
reason for changing the output to 5-step fringe images is that our dataset comprises high-frequency fringe images structured with 15 steps as its divisor. While predicting 15-step fringe images would be more appropriate for fair comparisons with other studies, we opted to reduce it to 5-step to align with the existing structure as closely as possible. Actually, there is not a significant difference between the wrapped phases constructed with 5 steps and 15 steps. Since the reference phase has the most substantial impact on the final output, the experiment proceeded with this modified output. Additionally, regarding the dataset used for the SIDO network, unlike our data, it contains a background, which is a reference plane. Through internal experiments, we determined that the presence or absence of a reference plane has a significant impact on the learning process, especially in the case of a single shot. Consequently, we conducted experiments by including information about the reference plane in the input.  

PSNet \cite{qi2023psnet} predicts N-step phase-shifted fringe images from a single fringe image. Like SIDO, a wrapped phase is predicted from predicted fringe images. To obtain an unwrapped phase, there should be more than two wrapped phases, so at least two networks are needed for 3D reconstruction. In this study, we trained two PSNet with two frequencies, $\lambda=\{15, 912\}$, respectively.

DLALNet \cite{zhang2022deep} is a network that performs phase unwrapping by calculating each corresponding warped phase with three fringe images with different frequencies as inputs. Unlike SIDO and PSNet, the numerator term and denominator term of the wrapped phase corresponding to each fringe image are predicted. In this study, it was used for comparison through an input consisting of $\lambda=\{15,114,912\}$. 

\begin{figure}[hbt!]
    \centering
    \begin{subfigure}{\textwidth}
        \includegraphics[width=\textwidth]{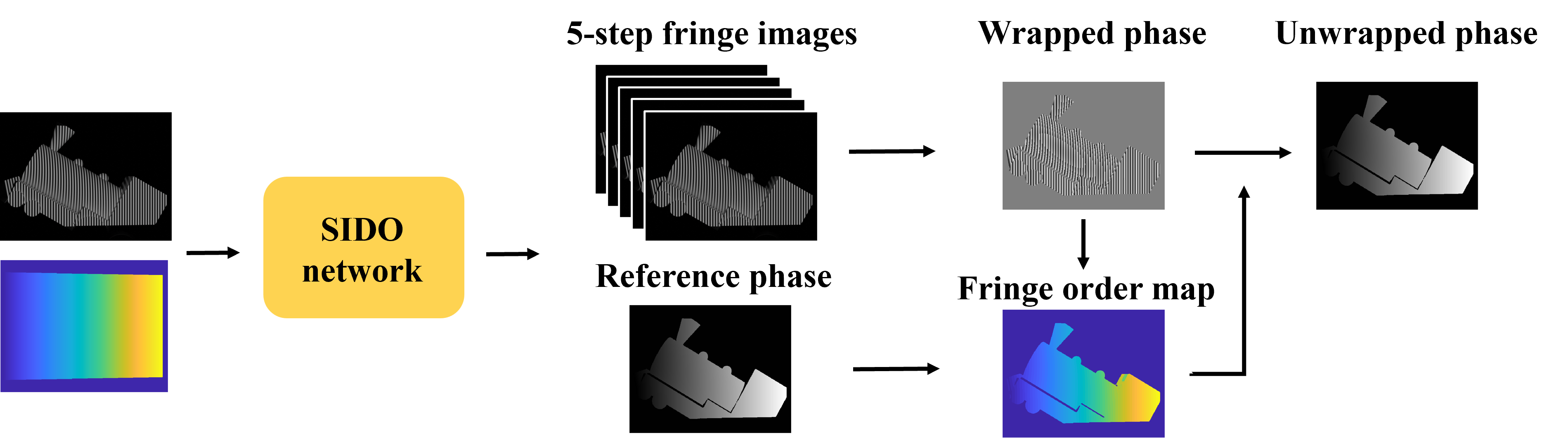}
        \caption{SIDO}
        \label{fig:sido}
    \end{subfigure}
    \hfill
    \begin{subfigure}{\textwidth}
        \includegraphics[width=\textwidth]{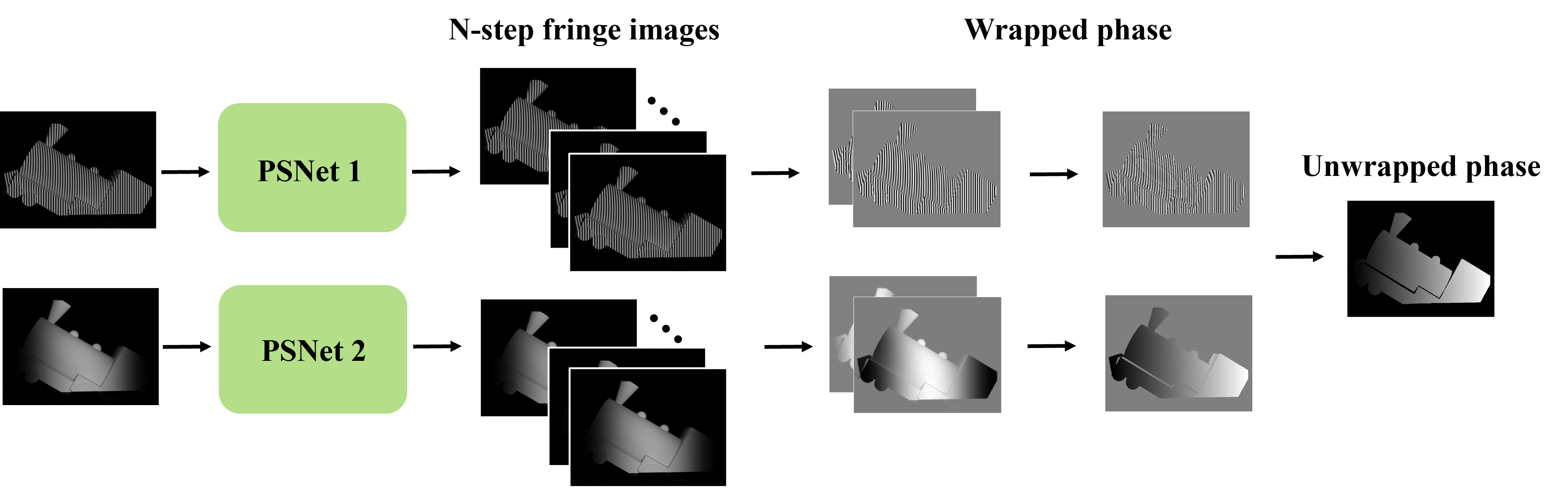}
        \caption{PSNet}
        \label{fig:psnet}
    \end{subfigure}
    \hfill
    \begin{subfigure}{\textwidth}
        \includegraphics[width=\textwidth]{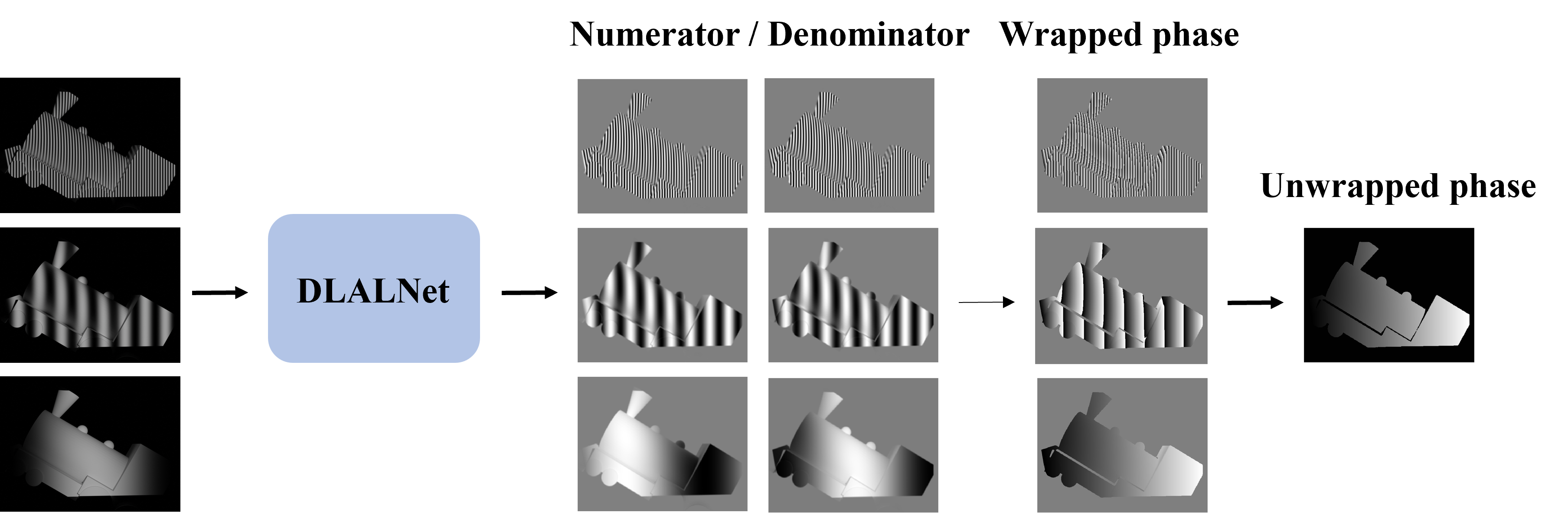}
        \caption{DLALNet}
        \label{fig:dlalnet}
    \end{subfigure}
        \caption{\textbf{Schematic diagram of baselines.}}
        \label{fig:baselines}
\end{figure}

Fig.~\ref{fig:baselines} shows the schematic diagram of baselines. The experimental setups primarily utilized configuration values as specified in each respective study, and training persisted until the loss was adequately saturated.

\subsection{Comparisons with other methods}
\begin{figure}[hbt!]
    \centering
    \includegraphics[width=\textwidth]{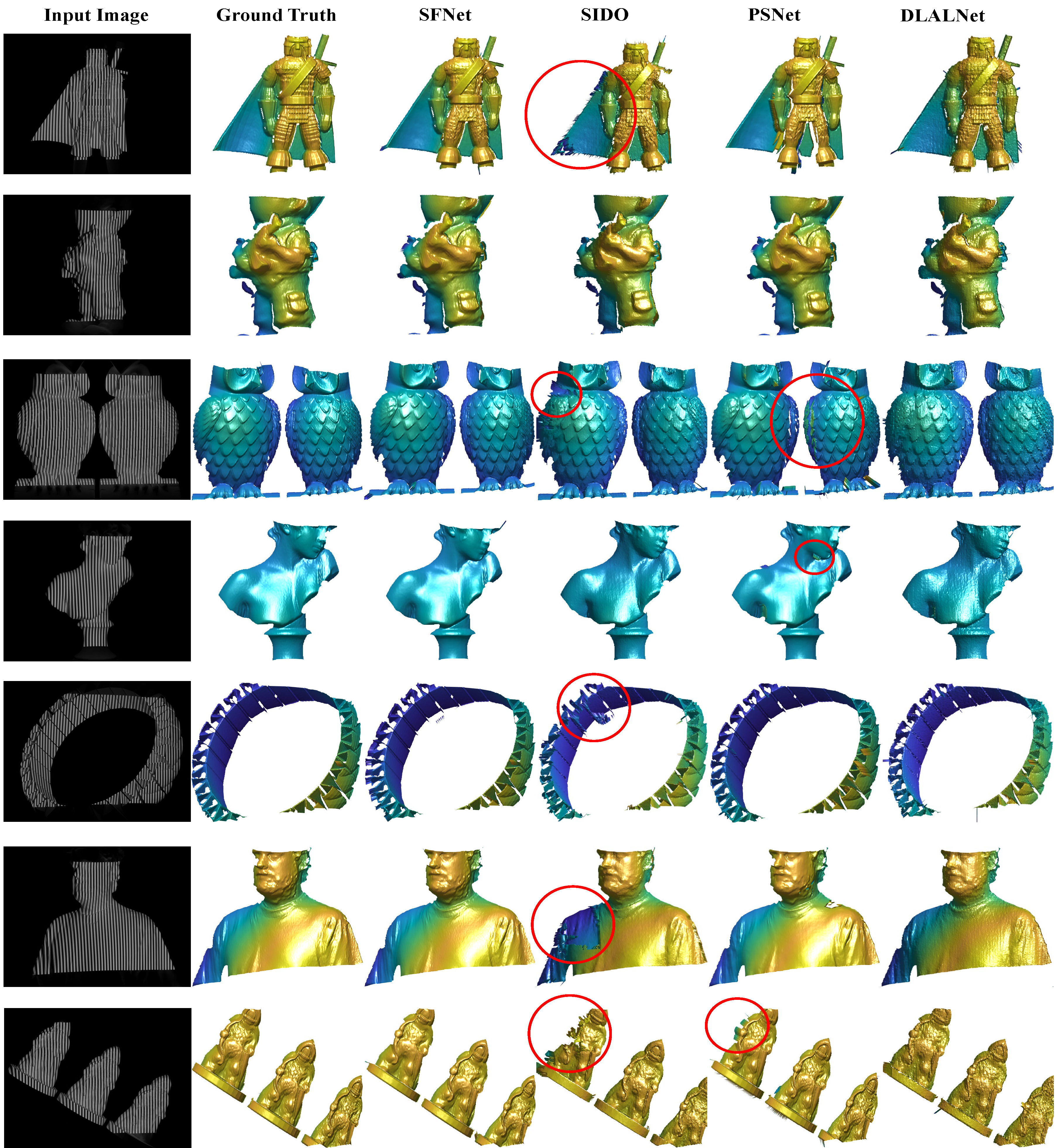}
    \caption{\textbf{Qualitative results.} Comparisons between the proposed method and other frameworks.}
    \label{fig:compare3D}
\end{figure}

\begin{figure}[hbt!]
    \centering
    \includegraphics[width=\textwidth]{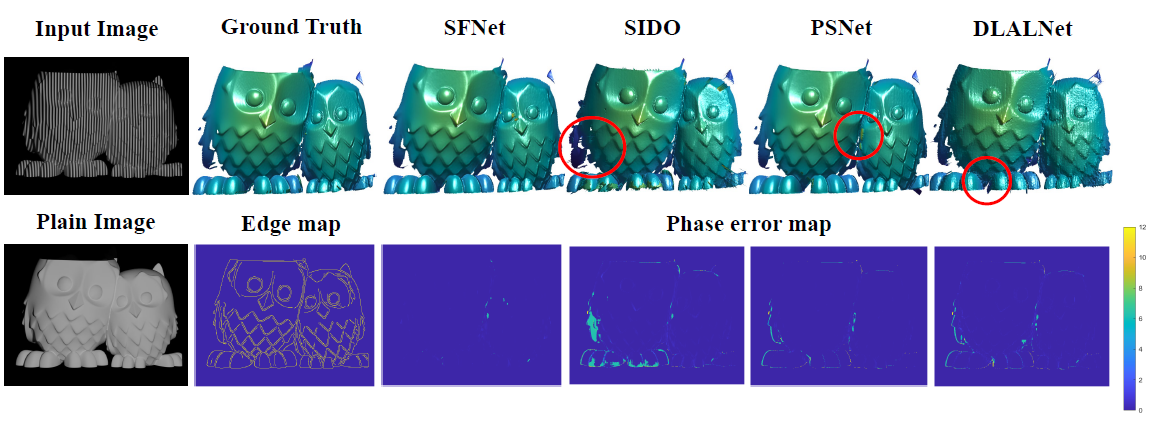}
    \caption{\textbf{Comparison of phase map error.} Input image and phase map errors between the ground truth and predicted output; The edge map is calculated from plain image (white pattern).}
    \label{fig:edge_phase}
\end{figure}

We compare the performance of SFNet with some baselines on our test dataset of 1800 image pairs. Each framework used in the comparison had its own frequency, $\lambda$, but since there was no dataset that encompassed all of them, we had no choice but to rely solely on our dataset for the comparison. As shown in Fig.~\ref{fig:compare3D}, it can be observed that all methods can roughly reconstruct the surface of target objects. However, our proposed method shows a better reconstruction capability, with smoother results, and fewer spike errors. On the other hand, the other methods present noticeable spike error and fringe order error especially along the edge of the target object, as shown in Fig.~\ref{fig:edge_phase}. For quantitative comparisons, we use evaluation metrics including mean absolute error (MAE) and root-mean-square error (RMSE) of unwrapped phase (rad) and depth value (mm). Table~\ref{table:comparison_uph_z} shows the quantitative results of the predicted results compared to the ground truth. 

First, SIDO network reveals the highest phase errors. This is because the information from a single fringe image is limited, leading to more substantial errors for reference phase, especially when dealing with surfaces featuring significant depth differences or complex geometry. The notable breakages in SIDO's results shown in Fig.~\ref{fig:compare3D} can be attributed to this limitation.
Next, in the case of PSNet, it consistently generates high-quality results, but spike errors arise occasionally. The predicted wrapped phase for each fringe image is generally accurate, with phase errors of $0.0063~rad$ and $0.0190~rad$ for high and low frequencies, respectively, indicating minor deviations. The emergence of undesired artifacts is attributed to employing two wrapped phases for phase unwrapping, particularly using the unit frequency wrapped phase. In contrast to the ground truth, which leverages three frequencies for more precise phase retrieval, PSNet utilizes only two frequencies, resulting in unwrapping errors as the quality of the reference phase decreases. While incorporating an additional frequency in the fringe image could address this issue, our current study concentrates on scenarios involving only two images.
Finally, in the case of DLALNet, there was generally effective restoration of the overall shape of objects; however, it was observed that wave artifacts occurred instead of smooth surfaces. This phenomenon is particularly attributed to small errors in the high-frequency wrapped phase, as the model's ability to correct the wrapped phase is diminished due to the use of a simple mean squared error (MSE) as the loss function. Consequently, such outcomes arise from this limitation.
In contrast, SFNet stands out for achieving a more accurate restoration of 3D surfaces compared to the previous results. It is remarkable for minimizing spike errors and obtaining smooth surfaces even for more complex shapes.

Fig.~\ref{fig:edge_phase} shows the phase map error of predicted results. Conventional phase unwrapping methods typically exhibit phase errors along the edges of objects. This trend persists in phase retrieval methods employing deep learning, where errors are primarily concentrated near the object's edges.  However, compared to other frameworks, it can be seen that the proposed method demonstrates notable accuracy not only across the entire area but also specifically near the object's edges.

\begin{table}
    \begin{center}
        \begin{tabular}{c | c c c c}
             \hline
                Method & MAE (rad) & RMSE (rad) & MAE (mm) & RMSE (mm)  \\
            \hline
                SFNet & \textbf{0.0527} & \textbf{ 0.6543} & \textbf{0.079} & \textbf{1.2359}  \\
                SIDO & 0.361 & 1.2945 & 1.6966 & 5.5933 \\
                PSNet & 0.0754 & 0.6759 & 0.1600 & 1.8549 \\
                DLALNet & 0.0908 & 1.3113 & 0.1438 & 1.8469\\
            \hline
        \end{tabular}
    \caption{Comparison of different methods using the same test dataset.}
    \label{table:comparison_uph_z}
    \end{center}
\end{table}


\subsection{Ablation studies}

\begin{figure}
    \centering
    \begin{subfigure}{0.9\textwidth}
        \includegraphics[width=\textwidth]{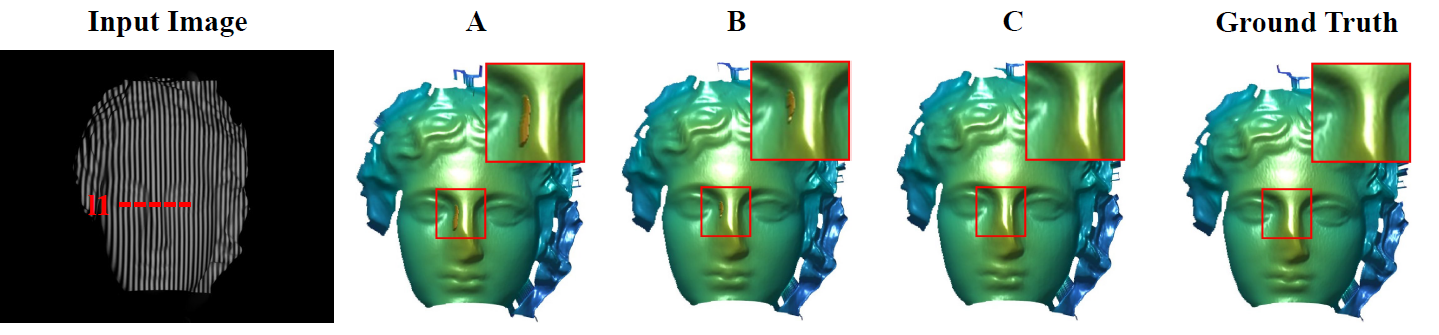}
        \caption{}
        \label{fig:3Dplot_1}
    \end{subfigure}
    \hfill
    \begin{subfigure}{0.9\textwidth}
        \includegraphics[width=\textwidth]{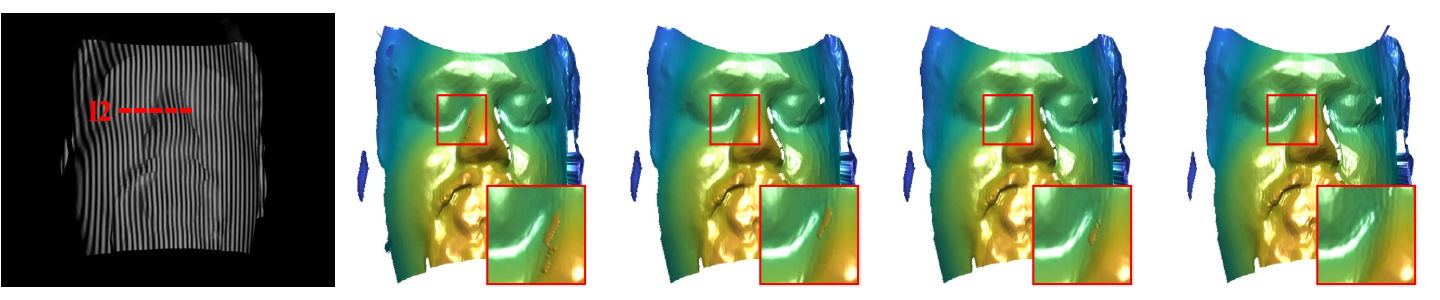}
        \caption{}
        \label{fig:3Dplot_2}
    \end{subfigure}
    \hfill
    \begin{subfigure}{0.9\textwidth}
        \includegraphics[width=\textwidth]{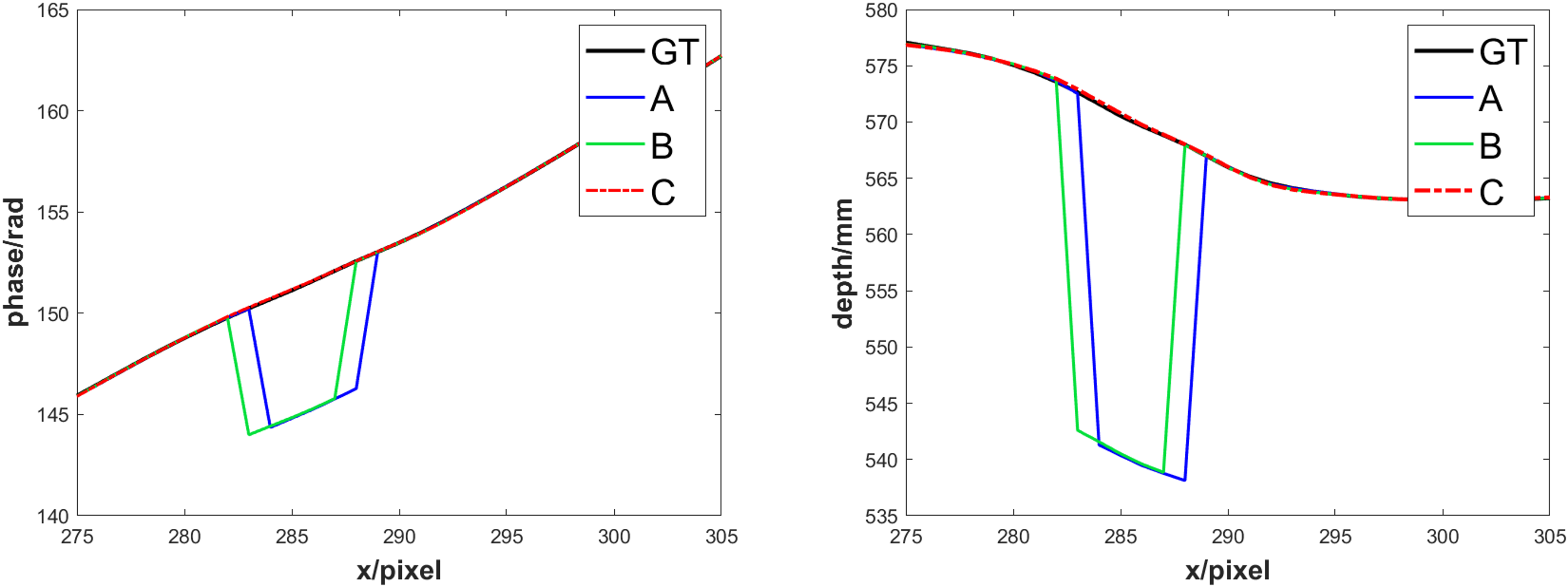}
        \caption{}
        \label{fig:abl_uph1}
    \end{subfigure}
    \hfill
    \begin{subfigure}{0.9\textwidth}
        \includegraphics[width=\textwidth]{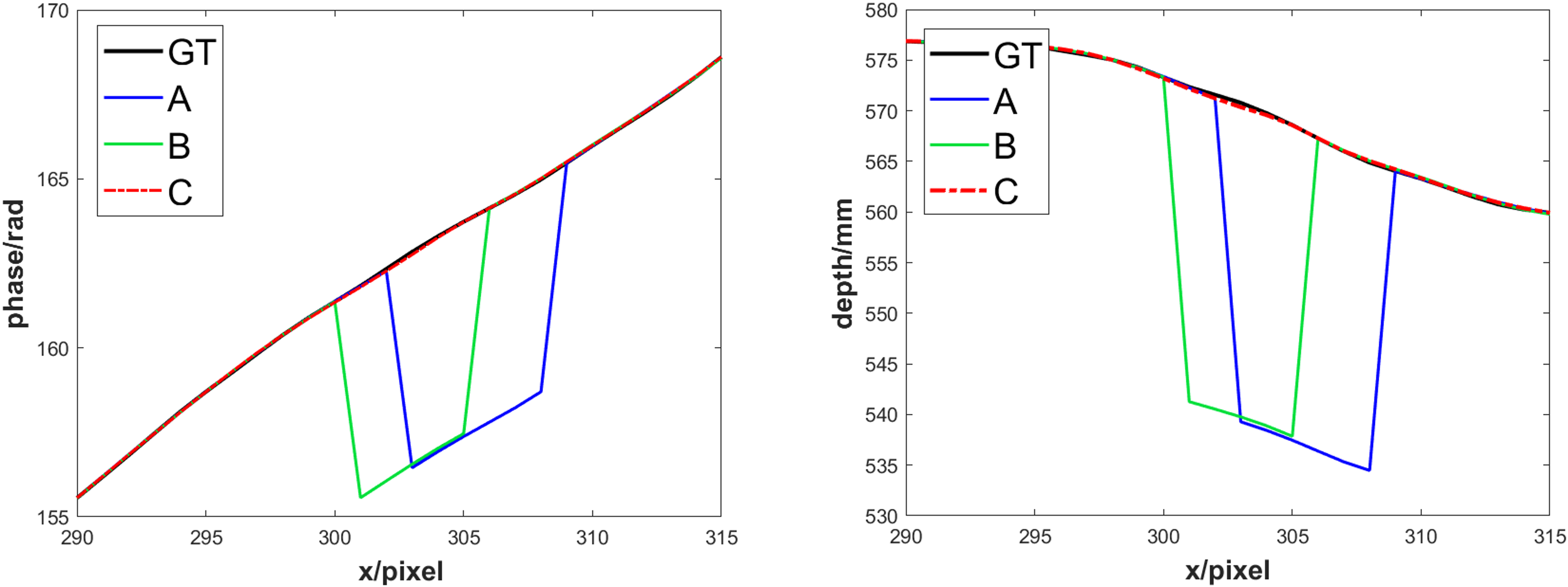}
        \caption{}
        \label{fig:abl_uph2}
    \end{subfigure}
        \caption{\textbf{Ablation study.} (a) - (b) Input fringe image (high-frequency) and corresponding 3D outputs of ablation study. (c) - (d) Cross-lines of the unwrapped phase (left) and the depth value (right) marked in the image in (a) and (b) with l1 and l2, respectively.}
        \label{fig:ablationstudy3D}
\end{figure}

\subsubsection{Decomposing the module of SFNet}

\begin{table}
    \centering
    \resizebox{\textwidth}{!}{\begin{tabular}{c | c c | c c c c}
             \hline
                 & Feature mixing & Our Loss (Eq.~\ref{equation:totalloss}) & MAE (rad) & RMSE (rad) & MAE (mm) & RMSE (mm) \\
            \hline
                A & & & 0.0744 & 0.8241 & 0.1589 & 1.8864 \\
                B & \checkmark & & 0.0546 & 0.6722 & 0.0865 & 1.2899 \\
                C & \checkmark & \checkmark & 0.0527 & 0.6543 & 0.079 & 1.2359 \\
            \hline
        \end{tabular}}
    \caption{Comparison of the results for decomposing the module of SFNet.}
    \label{table:ablation_uph_z}
\end{table}

In this part, we conducted the ablation experiments to prove the validity of the design of the SFNet. 
First, we decomposed SFNet into a few of sub-parts. The Fig.~\ref{fig:ablationstudy3D} and Table~\ref{table:ablation_uph_z} show the reconstruction results of the experiments. As shown in Table~\ref{table:ablation_uph_z}, the case of the method A showed the worst results This is because the reference phase of the proposed method is not a direct result from the corresponding fringe image, but is corrected through phases of another frequency. Therefore, the cases where richer information could be obtained through feature mixing showed better results. Also, the spike error occasionally occurs except in our method C. This is due to the existence of phase consistency and geometric constraint loss.

\subsubsection{Varying the number of input images}

\begin{table}
    \begin{center}
        \begin{tabular}{c | c c c c}
             \hline
                 Fringe images & MAE (rad) & RMSE (rad) & MAE (mm) & RMSE (mm) \\
            \hline
                2 (1 + 1) & 0.0527 & 0.6543 & 0.0790 & 1.2359 \\
                4 (2 + 2) & 0.0482 & 0.5581 & 0.0571 & 1.1621 \\
                6 (3 + 3) & \textbf{0.0441} & \textbf{0.5446} & \textbf{0.0477} & \textbf{1.1042} \\
            \hline
        \end{tabular}
    \caption{Comparison of results for varying the number of input images. }
    \label{table:ablation_study_2}
    \end{center}
\end{table}

Additionally, we conducted experiments by changing the number of fringe images used as input. Originally, the inputs consisted of one high-frequency and one low-frequency fringe image each. We modified this to use 2-step and 3-step fringe images for training. For the images at steps not present in our dataset, we captured additional images to include in the experiments.

Table~\ref{table:ablation_study_2} shows the results of the experiment. 
We observed that as the number of frames increased, both phase and depth errors diminished. However, when compared to previous experiments, the magnitude of the differences resulting from the increase in the number of frames was not significantly large. This suggests that the structure of our framework is sufficiently robust so is not heavily influenced by the additional information.

\subsubsection{Changes in the weight of the loss function}

\begin{table}
    \begin{center}
        \begin{tabular}{c | c | c c c c}
        \cline{1-6}
        $\lambda_{consist}$ & $\lambda_{geo}$ & MAE (rad) & RMSE (rad) & MAE (mm) & RMSE (mm) \\ \cline{1-6}
        \multirow{3}{*}{1e-1}   & 1e-5 & 0.0541 & 0.6707 & 0.0797 & 1.2594 \\
                               & 1e-6 & 0.0524 & 0.6546 & 0.0799 & 1.3225 \\
                               & 1e-7 & 0.0552 & 0.6841 & 0.0811 & 1.3368 \\ \cline{1-6}
        \multirow{3}{*}{1e-2}  & 1e-5 & 0.0562 & 0.7082 & 0.0832 & 1.2875 \\
                               & 1e-6 & 0.0527 & \textbf{0.6543} & \textbf{0.0790} & \textbf{1.2359} \\ 
                               & 1e-7 & \textbf{0.0508} & 0.6557 & 0.0801 & 1.2567 \\ \cline{1-6}
        \multirow{3}{*}{1e-3} & 1e-5 & 0.0562 & 0.6976 & 0.0808 & 1.2580\\
                               & 1e-6 & 0.0547 & 0.6723 & 0.0829 & 1.2834 \\
                               & 1e-7 & 0.0598 & 0.7004 & 0.0851 & 1.2927 \\ \cline{1-6}     
        \end{tabular}
    \caption{Comparison of results based on changes in the weight of the loss function. The bolded values are the first best-performing values.} 
    \label{table:ablation_study_3}
    \end{center}
\end{table}

Lastly, we adjusted the weights of the loss function as described in Eq.~\ref{equation:totalloss}. The results of our experiments are presented in Table~\ref{table:ablation_study_3}. It was observed that the Geometric Consistency Loss exhibited a more pronounced sensitivity to weight variations compared to the Phase Consistency Loss. The Phase Consistency Loss was employed to align the high-frequency wrapped phase with the tendency of the absolute phase. Assuming sufficient progress in training, the corrected portions involve intricate details, leading to relatively small differences in errors. However, in the case of the Geometric Constraint Loss, higher weight values between adjacent pixels reduce the disparity among them, resulting in a blurred restored phase. Conversely, if the weight is too small, it has a limited impact on the training process, contributing to the observed disparities as mentioned above.

%% file: secs/7_conclusion.tex
\section{Conclusion} \label{sec:conclusion}
The objective of this study is to obtain a high-quality 3D surface in the FPP system by utilizing a reduced number of fringe images. In this paper, we proposed a deep neural network, SFNet, that enables high-speed, high-quality 3D shape measurement even with a small number of fringe images. Our network receives two fringe images as input, performs feature extraction, fuses feature maps, and then recovers them to the corresponding wrapped phase. For the experiment, a virtual FPP environment was built and we could get a large dataset using the 3D CAD models. In addition, more reliable phase retrievals were possible through phase consistency loss and geometric construction loss. We demonstrated the efficiency of the proposed method by comparing it with existing fringe-to-phase approaches. Additionally, through various ablation studies, we demonstrated the validity of our framework structure.